\documentclass[10pt,twocolumn,letterpaper]{article}

\usepackage{cvpr}              
\usepackage[accsupp]{axessibility} 

\usepackage[dvipsnames]{xcolor}

\definecolor{cvprblue}{rgb}{0.21,0.49,0.74}
\usepackage[pagebackref,breaklinks,colorlinks,citecolor=cvprblue]{hyperref}
\usepackage[exponent-product=\times, retain-unity-mantissa=false]{siunitx} 
\usepackage{multirow}
\usepackage{multicol}
\usepackage{csquotes}
\usepackage{tikz}
\def\paperID{14} 
\def\confName{CVPR}
\def\confYear{2024}

 \usepackage[normalem]{ulem}
\usepackage[usestackEOL]{stackengine}
\usepackage[firstpage=true,color=black,placement=top,scale=1,opacity=1,vshift=-1cm]{background}
\SetBgContents{
 \begin{minipage}{2.3\linewidth}
 \small
\textcopyright 2024 IEEE.  Personal use of this material is permitted.  Permission from IEEE must be obtained for all other uses, in any current or future media, including reprinting/republishing this material for advertising or promotional purposes, creating new collective works, for resale or redistribution to servers or lists, or reuse of any copyrighted component of this work in other works. A definitive version will be published in \textbf{2024 IEEE/CVF Conference on Computer Vision and Pattern Recognition Workshops (CV4MS)}
 \end{minipage}
 }
\title{Self-Supervised Learning with Generative Adversarial Networks for Electron Microscopy}

\author{Bashir Kazimi\textsuperscript{1}, Karina Ruzaeva\textsuperscript{1}, Stefan Sandfeld\textsuperscript{1,2}\\
{\textsuperscript{1}Forschungszentrum J\"ulich GmbH, J\"ulich, IAS-9, Germany}\\
{\textsuperscript{2}RWTH Aachen University, Aachen, Germany}\\
{\tt\small \{b.kazimi,k.ruzaeva,s.sandfeld\}@fz-juelich.de}
}

\begin{document}
\maketitle
\begin{abstract}
In this work, we explore the potential of self-supervised learning with Generative Adversarial Networks (GANs) for electron microscopy datasets. We show how self-supervised pretraining facilitates efficient fine-tuning for a spectrum of downstream tasks, including semantic segmentation, denoising, noise \& background removal, and super-resolution. 
Experimentation with varying model complexities and receptive field sizes reveals the remarkable phenomenon that fine-tuned  models of lower complexity consistently outperform more complex models with random weight initialization. 
We demonstrate the versatility of self-supervised pretraining across various downstream tasks in the context of electron microscopy, allowing faster convergence and better performance. We conclude that self-supervised pretraining serves as a powerful catalyst, being especially advantageous when limited annotated data are available and efficient scaling of computational cost is important.  
\end{abstract}  
\section{Introduction}
\label{sec:introduction}

\begin{figure}[h]
    \centering
    \includegraphics[width=\linewidth]{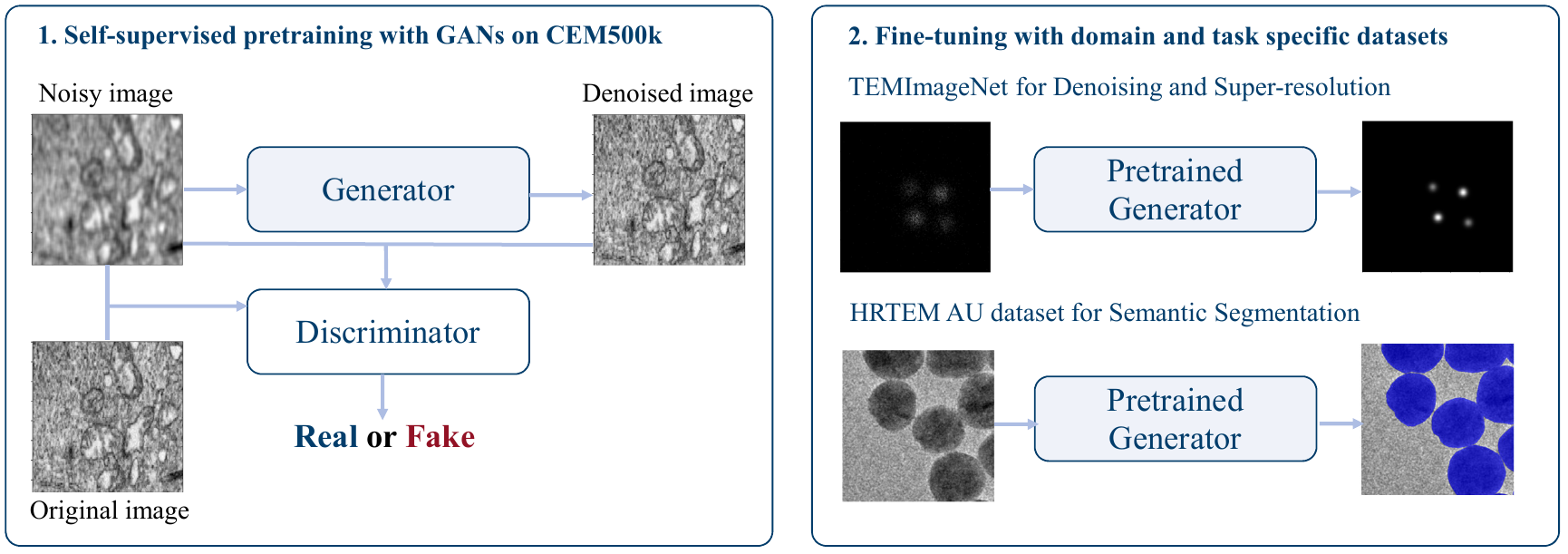}
    \caption{The proposed pretrainig pipeline, that includes GAN-based pretrainig on CEM500k dataset \cite{Conrad2021} followed by fine-tuning for downstream tasks: semantic segmentation of Gold nanoparticles \cite{Sytwu2022}, and super-resolution and denoising using the TEMImageNET dataset \cite{Lin2021}.
    }
    \label{fig:enter-label}
\end{figure}

Microscopy, a fundamental tool in scientific research for several centuries, encompasses various branches, including optical, electron, scanning probe, and X-ray microscopy \cite{leng2013materials}. Electron microscopy (EM) is a technique that uses a beam of accelerated electrons to obtain high-resolution images of biological as well as non-biological specimens. Applications of this technique exist across various scientific domains, including biology, materials science, nanotechnology, and physics. In the field of biology, EM has been used for studying a wide range of biological samples such as lungs, muscles, bones, or nerve tissue \cite{10.1016/s0074-7696(08)60048-0}. In materials science, it has been utilized for visualization of the growth and characterization of nano- and microstructures \cite{10.1021/jp5038415}, orientation mapping of semicrystalline polymers \cite{10.1016/j.micron.2016.05.008}, and the identification of crystal lattice defects \cite{Oh_Legros_Kiener_Dehm_2009}.

Imaging techniques and statistical analysis in EM have been instrumental in providing insights into the structure and properties of materials at various scales. Statistical analysis and classical machine learning methods have been used to analyze nanoparticles \cite{10.1021/acs.jpcc.0c07765,10.1021/acsnano.0c06809}, identify defects in metals \cite{Steinberger_Issa_Strobl_Imrich_Kiener_Sandfeld_2023, Zhang_Song_Oliveros_Fraczkiewicz_Legros_Sandfeld_2022}, and enhancing the quality of superresolution results in correlative tomography \cite{10.1073/pnas.1704908114}. These methods, however, have shortcomings, such as limited resolution, time-consuming sample preparation, and the need for expert interpretation of results. 

Deep learning (DL) and computer vision have been increasingly employed to address these limitations, enhance the capabilities of EM, and overcome the limitations of classical imaging and analysis methods by providing automated analysis, improved resolution, and enhanced interpretation of complex data:
DL enables the extraction of valuable information from large datasets and offers new opportunities for quantitative image analysis in EM \cite{10.15252/msb.20156651}. It has been used for analyzing nanoparticles in TEM images \cite{Sytwu2022}, denoising TEM images \cite{10.1017/s1431927621012678}, identifying clean graphene areas \cite{sadre2021deep}, automatically segmenting and tracking of crystalline defects, \cite{Ruzaeva_Govind_Legros_Sandfeld_2023, Govind_Oliveros_Dlouhy_Legros_Sandfeld_2024}, decoding crystallography from high-resolution electron imaging and diffraction datasets \cite{10.1126/sciadv.aaw1949}, understanding important features of DL models for segmentation of high-resolution transmission electron microscopy (TEM) images \cite{Horwath2020}, and segmentation in large-scale cellular EM \cite{ASWATH2023102920}, focused ion-beam scanning EM (FIB-SEM) \cite{Khadangi2021} and high-resolution TEM data \cite{Groschner2021}.

Conventional DL methods, such as convolutional neural networks, require large annotated datasets to be able to learn and generalize well on unseen examples. Manual annotation of datasets, especially EM images, is a time-consuming and labor-intensive task. To alleviate this, techniques such as transfer learning and self-supervised learning can be used. 
These techniques offer significant advantages for a range of computer vision tasks by enabling models to obtain general features from extensive datasets, facilitating knowledge transfer to specific tasks with limited labeled data. These approaches significantly reduce annotation costs, mitigate the problem of data scarcity, and strongly enhance generalization to unseen scenarios. Pretrained models exhibit faster convergence during fine-tuning, possess broader applicability across tasks, and provide resource-efficient solutions. The robust representations acquired through self-supervised learning contribute to improved performance in real-world scenarios, establishing it as an essential strategy in computer vision tasks across diverse domains \cite{he2020momentum,10.48550/arxiv.2102.01530,10.1155/2022/3264367,10.18383/j.tom.2016.00211}. 

Self-supervised learning aims to learn representations from the data itself without explicit manual supervision. It can be utilized to pretrain a model on a large amount of unlabeled data, allowing it to learn general features and representations from the data. These learned representations can then be transferred and fine-tuned for a specific task, effectively leveraging the knowledge gained from the self-supervised pretraining to improve performance on the target task. Pretraining models on these tasks with unlabeled data and using the pretrained weights to fine-tune models on supervised tasks with limited annotations help improve model performance and reach faster convergence \cite{pathak2016context,chen2020simple,he2020momentum}. The first step in self-supervised learning (pretraining on unlabeled data) is called the pretext. The second step (fine-tuning the pretrained models on annotated data) is called downstream.

The main goal of this research is to use Generative Adversarial Networks (GANs) in a self-supervised learning framework (Fig.~\ref{fig:enter-label}) to pretrain models on large unlabeled EM datasets and use the weights to fine-tune DL models for various supervised downstream tasks such as semantic segmentation of nanoparticles, denoising, super-resolution, and noise \& background removal in high-resolution TEM images. We show that such pretraining generalizes well and results in faster convergence and improved performance for different kinds of supervised tasks in EM with limited annotated data. Additionally, pretraining alleviates the need for training complex network architectures and expensive hyperparameter optimization. As a benchmark, results are compared with the work of \citeauthor{Sytwu2022}, which investigates the impact of receptive field size on the performance of DL models for semantic segmentation of nanoparticles in TEM images. Our results show that pretraining on unlabeled data leads to an improved performance regardless of the receptive field size or network architecture. More specifically, with fine-tuning, simple and smaller models achieve at least similar, often even better performance compared to larger, more complex models with randomly initialized weights.

This work makes the following scientific contributions: (i) demonstrating the substantial performance and convergence improvements in EM tasks through self-supervised (GAN-based) pretraining on unlabeled images; (ii) highlighting the generalization benefits and reduced dependency on hyperparameter optimization across different network architectures and receptive field sizes; (iii) a versatile framework for fine-tuning DL models on various EM tasks is introduced, including semantic segmentation, denoising, noise and background removal, and super-resolution.

\section{Related Work}
\label{sec:relatedwork}

A commonly used approach in deep learning is pretraining models on large labeled datasets and fine-tuning on smaller datasets with limited annotations \cite{ridnik2021imagenet,zhang2020transfer,marmanis2015deep}. Such an approach usually performs well when the source data for pretraining is from a domain similar to the one from which the target data was obtained. In domains where large labeled datasets are scarce but an abundance of unlabeled datasets is available, self-supervised learning proves to be effective. Self-supervised learning leverages unlabeled datasets for pretraining, and the learned knowledge is then transferred to supervised downstream tasks with labeled data. Examples of successful self-supervised learning methods include contrastive learning, jigsaw puzzles, autoencoders, masked image modeling, and generative-based methods. SimCLR \cite{chen2020simple} is a prominent self-supervised learning method that has demonstrated significant advancements in self-supervised learning on large-scale benchmarks such as ImageNet. It leverages a contrastive pretraining objective, which involves maximizing agreement between differently augmented views of the same data point while minimizing agreement with views from other data points. This approach has been shown to learn semantically meaningful representations from unlabeled data, making it a powerful method for self-supervised learning. Momentum Contrast (MoCo) \cite{he2020momentum} is another well-known technique that leverages contrastive learning for unsupervised visual representation learning. It has been widely applied in various domains, including remote sensing scene classification \cite{10.1038/s41598-022-27313-5}, chest X-ray model pretraining \cite{10.48550/arxiv.2010.05352}, hand shape estimation \cite{zimmermann2021contrastive}, and speaker embedding \cite{10.48550/arxiv.2001.01986}. The method has also been compared with other self-supervised learning techniques, demonstrating its effectiveness in learning representations from images and its potential for various downstream tasks \cite{10.1609/aaai.v36i3.20248}. ~\citeauthor{Conrad2021} used this method to pretrain DL models on cellular EM images.

\begin{figure}[!h]
    \centering
    \includegraphics[width=.45\columnwidth]{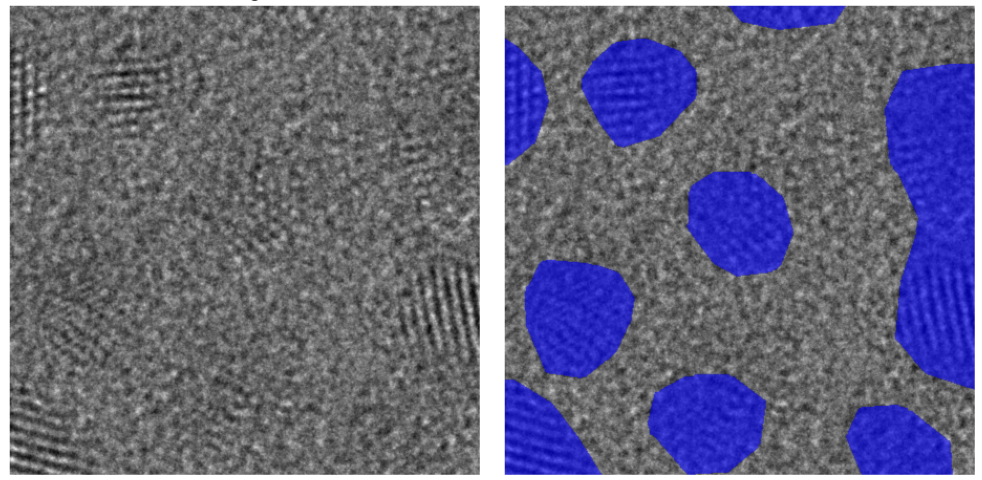}
    \hfill
    \includegraphics[width=.45\columnwidth]{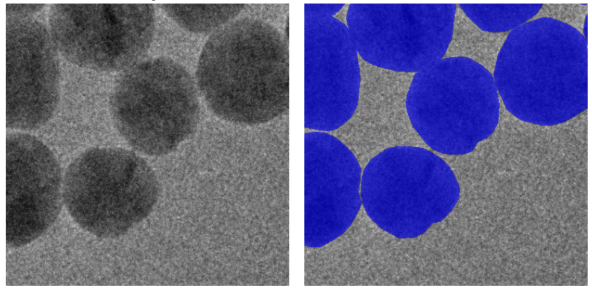}
    \caption{High- (left) and low-resolution (right) TEM image dataset of \SI{2.2}{\nano\meter}  and \SI{20}{\nano\meter} Au nanoparticles and their ground truth segmentations. The more ordered structures are the nanoparticles, and the noisy regions are the amorphous matrix.}
    \label{fig:auex}
\end{figure}

\begin{figure}[!ht]
    \centering
    \includegraphics[width=.23\columnwidth]{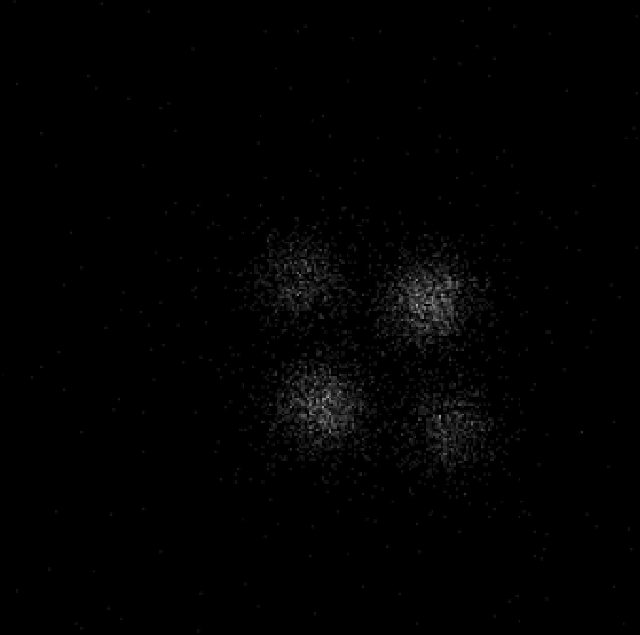}
    \includegraphics[width=.23\columnwidth]{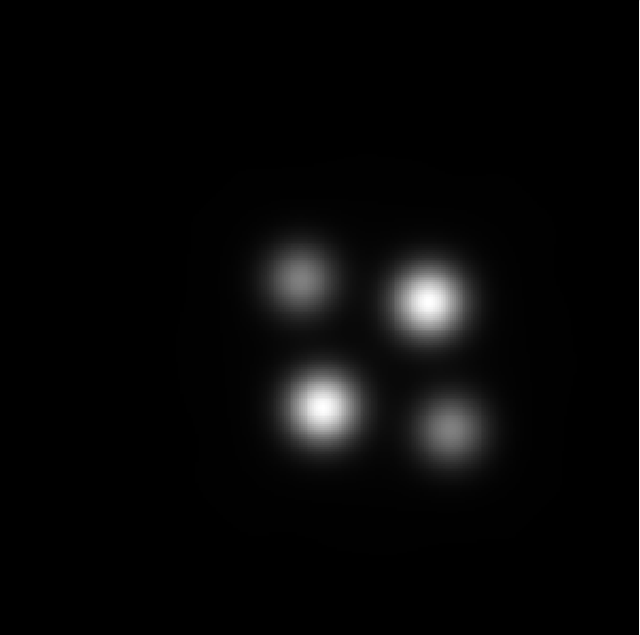}
    \includegraphics[width=.23\columnwidth]{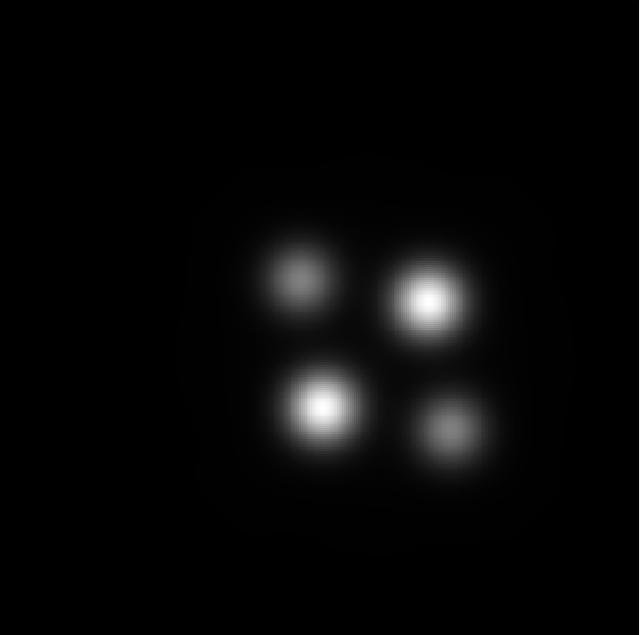}
    \includegraphics[width=.23\columnwidth]{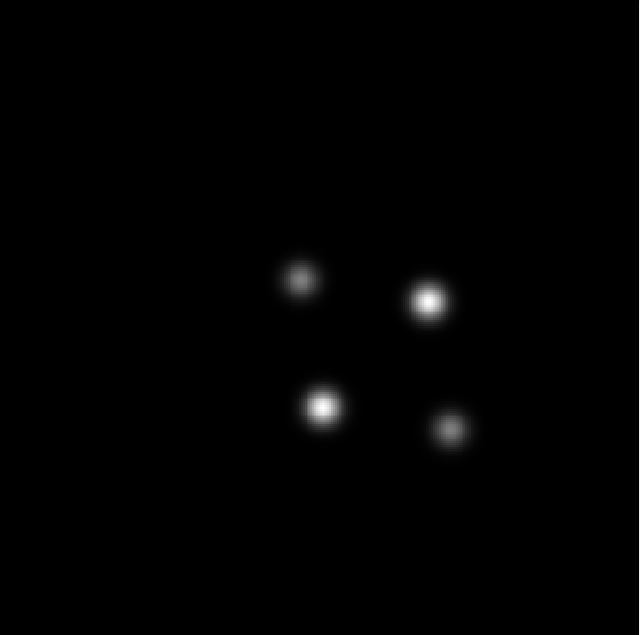}
    \caption{The example TEMImageNet image and corresponding ground truth labels. Left to right: original image, noise reduction, denoising \& background removal, and super-resolution}
    \label{fig:tinex}
\end{figure}

Masked image modeling, another self-supervised learning method, involves training a model to predict the original content of an image from a corrupted or masked version \cite{xie2022simmim}. This approach has been applied in various domains, including medical imaging and spectroscopic data identification. \citeauthor{10.48550/arxiv.2211.00313} proposed RGMIM (Region-Guided Masked Image Modeling) for COVID-19 detection, showcasing the potential of masked image modeling in medical imaging applications. Furthermore, \citeauthor{10.48550/arxiv.2211.08887} highlighted the success of masked image modeling in self-supervised learning, demonstrating its ability to alleviate data-hungry issues and achieve competitive results. \citeauthor{Caron_2021_ICCV} show the prominence of self-supervised learning on various tasks using vision transforms. These examples underscore the significance of masked image modeling in learning robust representations and its applicability across diverse domains. 

Using GANs for self-supervised pretraining is also very effective. \citeauthor{chen2019self} uses a GAN-based model to pretrain a model that learns image rotation. \citeauthor{guo2021self} uses GAN-based pretraining for learning image similarity in remote sensing images. Other notable research in this area includes latent transformation detection \cite{patel2021lt}, GAN-based image colorization for self-supervised visual feature learning \cite{treneska2022gan}, and self-supervised learning for semantic segmentation of archaeological monuments \cite{kazimi2023self}. 

Recently, pretraining methodologies have been explored in the domain of EM. In particular, the microstructure segmentation with DL encoders pretrained on a large microscopy dataset \cite{Stuckner2022}, classification of scanning electron microscope images of pharmaceutical excipients using deep convolutional neural networks with transfer learning \cite{Iwata2022} and the unsupervised pretraining, the Momentum Contrast (MoCoV2) algorithm \cite{Chen2020} was used by \citeauthor{Conrad2021}.

In this paper, we explore the application of self-supervised pretraining based on GANs, specifically the Pix2Pix architecture \cite{Isola2017}, for EM images. The GAN model is pretrained on a large unlabeled Cellular Electron Microscopy (CEM) dataset called CEM500K \cite{Conrad2021}. The pretrained generator model can be fine-tuned on a wide range of downstream tasks in EM, including semantic segmentation of nanoparticles in TEMs, denoising, noise \& background removal, and super-resolution. We show that such a pretraining approach leads to faster convergence and higher predictive power on all of the mentioned tasks with limited annotated datasets. We also find that fine-tuning with pretrained weights helps smaller and architecturally simpler models achieve similar or even higher scores compared to training with random weight initialization. 

\section{Materials and Methods}
\label{sec:materialsandmethods}

\subsection{Datsets}

\subsubsection{CEM500K}
CEM500K \cite{Conrad2021} is a large-scale, heterogeneous, unlabeled cellular EM image dataset developed for DL applications. The dataset is curated from experiments and various publicly available sources, encompassing 2D and 3D cellular EM images with diverse imaging modalities, sample preparation protocols, resolutions, and cell types. It includes examples from reconstructed FIB-SEM volumes, transmission EM (TEM) images, and EM image volumes and 2D images from various sources.

\subsubsection{HRTEM Au Dataset}
The Gold nanoparticle image dataset consists of high- and low-resolution TEM images of Gold (Au) nanoparticles with varying sizes (\SI{2.2}{\nano\meter}, \SI{5}{\nano\meter}, \SI{10}{\nano\meter}, and \SI{20}{\nano\meter}) and different surface ligands, i.e., citrate (for \SI{2.2}{\nano\meter}) and tannic acid. The images were acquired using an aberration-corrected TEAM 0.5 TEM for 2.2, 5, and \SI{10}{\nano\meter} nanoparticles, while low-resolution images of \SI{20}{\nano\meter} nanoparticles were obtained with a non-aberration-corrected TitanX TEM. 
The nanoparticles have a different crystalline structure than the embedding matrix (which is amorphous). This is
the reason why the atomic arrangements look different in TEM. An important task in materials science is to segment such nanoparticles.
The dataset was manually segmented and labeled, followed by preprocessing steps such as outlier removal and image standardization. To optimize memory usage during training, images were divided into 512 × 512-pixel patches, excluding patches consisting solely of amorphous background to address potential class imbalance issues \cite{Sytwu2022}. In this paper, we will refer to the datasets as
\enquote{Au2.2nm}, \enquote{Au5nm}, \enquote{Au10nm},\enquote{Au20nm}.
Additionally, the dataset of \SI{5}{\nano\meter} Au nanoparticles \cite{Groschner2021} was included in our experiments and is referred to as \enquote{Au5nmV1}. Examples of low and high-resolution TEM images of Gold nanoparticles and the corresponding ground truth segmentation annotations are shown in Figure~\ref{fig:auex}.

\subsubsection{TemImageNET}
TEMImageNet is an open-source atomic-scale scanning transmission electron microscopy (ADF-STEM) image dataset. The dataset includes ten types of ground truth labels for training and validating DL models for tasks such as segmentation, super-resolution, background subtraction, denoising, and localization.
The dataset comprises simulated ADF-STEM images of eight materials projected along multiple orientations with diverse atomic structures and crystallographic orientations. To replicate real-world experimental conditions, the images are augmented with realistic scan and Poisson noise, along with randomized linear and nonlinear low-frequency background patterns \cite{Lin2021}. The example simulated image and the ground truth labels are shown in Figure \ref{fig:tinex}. The number of images in each dataset for training, validation and testing is given in Table~\ref{data-table}.

\begin{table}[t]
\caption{Number of images for training validation and testing of the datasets, used in the experiments.}
\label{data-table}
\vskip 0.15in
\begin{center}
\begin{small}
\begin{sc}
\resizebox{\linewidth}{!}{
\begin{tabular}{lcccr}
\toprule
Dataset & Train & Validation & Test \\
\midrule
CEM500K &50k/100k/200k & 5000 & 5000\\
Au10nm & 660 & 220 &220 \\
Au5nmV1 & 144 & 48 & 48\\
Au5nm  & 1044 & 348 & 348\\
Au20nm  & 660 &  220 & 220\\
Au2.2nm     & 1740 & 580 &580\\
TEMImageNet     & 10377 & 1832 &2155\\

\bottomrule
\end{tabular}}
\end{sc}
\end{small}
\end{center}
\vskip -0.1in
\end{table}

\subsection{Pretraining Method}
\label{sec:pretrainingmethod}
We employ a GAN-based approach for pretraining on unlabeled data. GANs are originally designed for generating new data samples that resemble a given dataset. The GAN architecture involves two neural networks: a generator and a discriminator. These are trained simultaneously through adversarial training. The generator takes random noise as input and generates synthetic data samples that are indistinguishable from real data. The discriminator evaluates the real and generated data and distinguishes them from each other. Both networks are simultaneously trained in an adversarial fashion: the generator tries to improve its ability to generate realistic data to fool the discriminator. The discriminator, in turn, strives to become better at distinguishing between real and generated samples \cite{goodfellow2014generative}. 

Conditional Generative Adversarial Networks (cGANs) are an extension of the traditional GAN framework, where the generator is conditioned on additional information, typically in the form of class labels or other auxiliary data. The key idea is to guide the generation process based on specific conditions, allowing a more controlled and targeted generation of samples. In a cGAN, both the generator and the discriminator receive additional input information (conditioning) alongside the random noise for the generator and real/fake labels for the discriminator. The conditioning information could be anything relevant to the desired output, e.g., class labels, attributes, or other types of data. Conditional GANs have been used for various tasks, including image-to-image translation, image synthesis with specific attributes, and generating samples from certain classes. They provide a way to control and manipulate the characteristics of the generated data by incorporating additional information during the training process \cite{10.48550/arxiv.1411.1784}. The same loss function for cGANs as mentioned in \cite{Isola2017} is used.

In this paper, we use the cGAN model called Pix2Pix \cite{Isola2017} for pretraining on the unlabeled CEM500K dataset of EM images. The images are fed to the generator with added noise, and the goal is to generate output images that are indistinguishable from the original ones. The trained generator model can then be fine-tuned on supervised downstream tasks explained in Section \ref{sec:downstream}. To investigate the influence of the size of unlabeled datasets used for pretraining on the results of fine-tuning in the downstream tasks, we run pretraining experiments with different numbers of examples from the CEM500K dataset: 50K, 100K, and 200K examples.

\subsection{Downstream Tasks}
\label{sec:downstream}

The pretrained generator model explained in the previous section can be fine-tuned on a wide range of downstream tasks. As case studies in this paper, we have selected tasks including semantic segmentation of nanoparticles, denoising, noise \& background removal, and super-resolution in high-resolution TEM images. 

Semantic segmentation involves labeling each pixel of an image with a corresponding class of what is being represented. In materials and biological sciences, this imaging task plays a crucial role in analyzing and understanding complex microstructural and elemental features within the images obtained. \citeauthor{Sytwu2022} conducted experiments on semantic segmentation of Gold nanoparticles of different sizes and resolutions using the U-Net model \cite{ronneberger2015u}. They used a U-Net with different numbers of residual blocks and different receptive field sizes to study the influence of model complexity and receptive field size on the performance of the model. Their findings show that increasing the receptive field increases model performance in high-resolution TEM images. They also conclude that as the model complexity increases, there is a corresponding improvement in performance and prediction confidence. Here, the focus is on an investigation of the influence of pretraining on the performance of DL models in semantic segmentation. Therefore, we selected the same network as that used by \citeauthor{Sytwu2022}, i.e., the U-Net, and used it as the generator for pretraining on unlabeled data in the previous step. We then fine-tuned it for semantic segmentation on the same dataset, i.e., high-resolution TEM images of Gold nanoparticles. The results were compared to training the same model with randomly initialized weights. These experiments were conducted with different complexities and receptive field sizes. We additionally used a more complex High-Resolution Network (HRNet) \cite{SunXLW19} as the generator and fine-tuned it on this dataset for comparison. 

Furthermore, atom segmentation, localization, noise reduction, and deblurring are crucial tasks in atomic-resolution scanning transmission electron microscopy (STEM). The images captured at the atomic scale often suffer from noise, which can obscure subtle details and compromise the accuracy of atom segmentation and localization \cite{Lin2021}. Denoising is important in refining these images, ensuring a clearer representation of the atomic structure. It enhances the level of detail in the images beyond the inherent resolution of the microscope, achieving higher spatial resolution, which allows researchers to discern finer structural features and better characterize atomic arrangements. By reducing noise and enhancing resolution, the method ensures a more accurate and robust analysis of atomic structures, even in challenging imaging conditions with variations in sample thickness. 
To show the generalizability of pretrained models in our work, we use the TEMImageNet data and do experiments on denoising, noise \& background removal, and super-resolution. We use the HRNet model and run the same experiments to compare the results of fine-tuning to those of training with random weight initialization. Details of experiments and results are outlined in Section \ref{sec:experimentsandresults}.

\begin{figure}[h]
    \centering
    \includegraphics{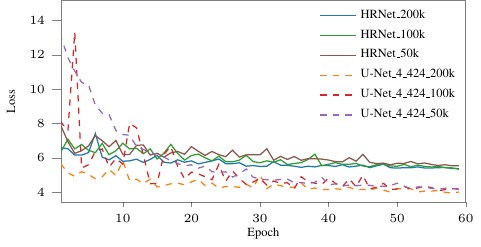}
    \caption{Validation $L_1$ loss for HRNet and U-Net\_4\_424 for three different dataset sizes.}
    \label{fig:ganplot}
\end{figure}

\begin{figure}
    \centering
    \includegraphics[width=.23\columnwidth]{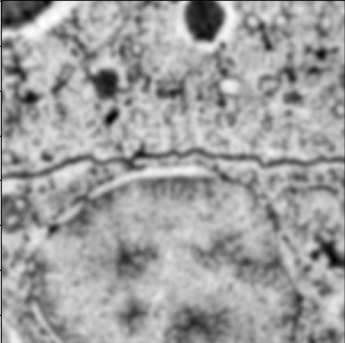}
    \includegraphics[width=.23\columnwidth]{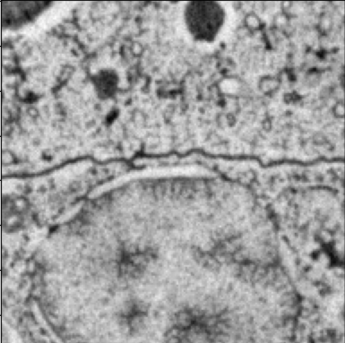}
    \includegraphics[width=.23\columnwidth]{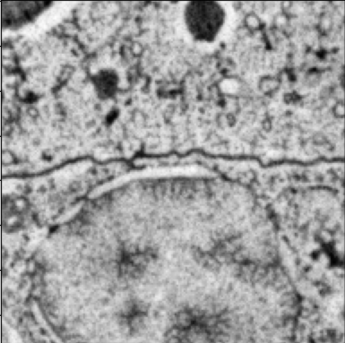}
    \includegraphics[width=.23\columnwidth]{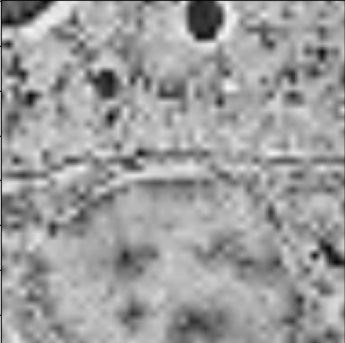}
    \caption{Left to write: input image with added noise, original image, and generated images by U-Net and HRNet.}
    \label{fig:generatorresults}
\end{figure}

\section{Experiments and Results}
\label{sec:experimentsandresults}

\subsection{Experiments on CEM500K and GANs} 
The CEM500K dataset was employed in the pretraining step where cGANs, based on the Pix2Pix 
model \cite{Isola2017}, were used. As generator, we used different U-Net architectures 
with varying numbers of residual blocks and receptive field sizes. Specifically, configurations 
with 2, 3, and 4 residual blocks were considered, each associated with different receptive 
field sizes. For two residual blocks, the receptive field sizes were 44, 84, and 116, for three blocks, 
we used 96, 176, and 240 as well as a receptive field of 200, 360, and 424 for four blocks, 
in line with the work presented in \cite{Sytwu2022}. For brevity, we refer to these U-Net variations as U-Net\_{B}\_{RF} from now on, where B refers to the number of residual blocks and RF refers to receptive field size.  Additionally, a more complex model 
architecture, HRNet \cite{wang2021}, was used to explore its effectiveness in comparison 
to U-Net variants and, in particular, to understand how far an increase in model complexity 
results in an increase in expressivity.

For training, random Gaussian noise, blurring, flipping, and rotations were applied to the 
images used as input to the generator. The generator was trained to predict images indistinguishable from the original ``clean'' images. The training for each model variation was conducted for \SI{60}{} epochs with a batch size of \SI{128}{}. Adam optimizer with a learning rate of \SI{2e-4}{} was used for optimization. In terms of training the whole GAN architecture, it was framed as the Least Square GAN (LSGAN), which adopts the least squares loss function for the discriminator and is more stable than regular GANs. LSGANs are able to generate higher quality images and perform more stably during the learning process \cite{mao2017least}. The generator was also trained using the $L_1$ loss, and the $\lambda$ in Equation \ref{eq:overallgan} was set to \SI{100}{}. 

Each model variation was trained with the same experimental setup for three different subsets of the CEM500K dataset consisting of \SI{50}{K}, \SI{100}{K}, and \SI{200}{K} images. This experiment was conducted to investigate the influence of the dataset size in pretraining on the model performance during fine-tuning on downstream tasks. 

The validation plots for the GAN pretraining with HRNet and the most complex U-Net variation are shown in Figure \ref{fig:ganplot}. For each of the models, as the dataset size increases, the validation loss decreases. Interestingly, the U-Net model shows better validation results compared to HRNet. This is also illustrated in the generated images in Figure \ref{fig:generatorresults}. We believe this is due to the skip connections from the initial residual blocks in the U-Net model to the corresponding upsampling blocks. The U-Net model has direct access to the information in the larger spatial resolutions and is more prone to memorizing, while the HRNet model actually learns the feature representations as it encodes the images into lower spatial resolutions with higher feature maps and then decodes it, without having direct access to the input features in the higher spatial resolution. Additionally, we find that the pretrained HRNet model fine-tuned on a variety of supervised tasks outperforms the U-Net model. For brevity, results for other experiments are included in the supplementary materials.


\subsection{Experiments on Semantic segmentation with Au datasets} 

The segmentation of nanoparticles in the Au datasets was approached using the same variations of the U-Net and the HRNet models that were pretrained on the unlabeled CEM500K dataset. All model variations were trained and systematically compared with respect to weight initialization with random weights and with the pretrained weights from the previous step.
All experiments were performed by training for 60 epochs on each dataset. As an objective function, the Binary Cross Entropy (BCE) loss was used and minimized by the Adam optimizer with a learning rate of \SI{2e-4}{}. The model's performance was evaluated using the dice score. The original image size is $512 \times 512$ pixels, but data augmentation techniques, including various types of noise, flips, rotations, resizing, and random cropping to $448 \times 448$ pixels, are applied.

\begin{figure}
    \centering
    \includegraphics[]{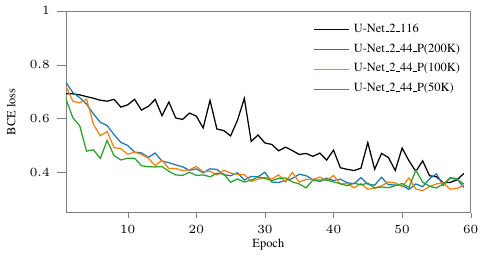}
    \vspace{-.4cm}
    \caption{Comparison of validation loss for bigger randomly initialized U-Net and smaller fine-tuned U-Nets. }
    \label{fig:seg1}
\end{figure}

\begin{figure}
    \centering
    \includegraphics[]{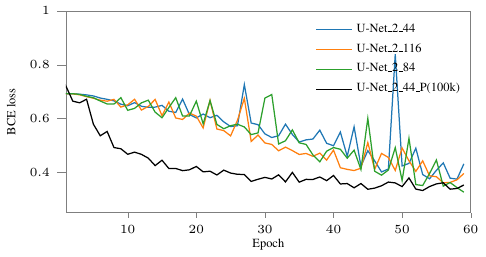}
    \vspace{-.4cm}
    \caption{Comparison of validation loss for smaller fine-tuned U-Net and randomly initialized U-Nets of varying complexity.}
    \label{fig:seg2}
\end{figure}

In the experiments conducted with different variations of the U-Net, we observe that randomly initialized models with a higher receptive field size perform better than those with a smaller receptive size. However, the fine-tuned models with smaller receptive field sizes initially not only outperform the same models with randomly initialized weights, but even outperform those with bigger receptive fields. Even though the randomly initialized model catches up as training progresses longer, the performance is not stable and the oscillations are large. This is illustrated in the validation plots for the Au5nmV1 dataset in Figure \ref{fig:seg1}. We also observe that the fine-tuned U-Net model (U-Net\_2\_44\_P(100K)) outperforms the randomly initialized models, as illustrated in Figure \ref{fig:seg2}. Suffixes P(50K), P(100K) and P(200K) indicate models that were pretrained on \SI{50}{K}, \SI{100}{K}, and \SI{200}{K} unlabeled images, respectively, in the first step.

\begin{figure}
    \centering
    \includegraphics[]{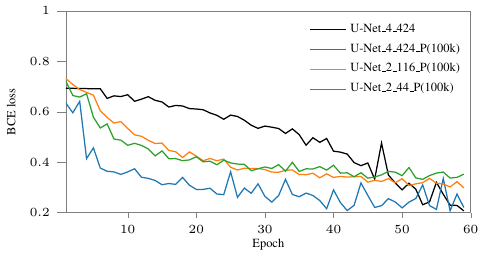}
    \vspace{-.4cm}
    \caption{Validation loss for bigger randomly initialized U-Net and fine-tuned U-Nets of varying size.}
    \label{fig:seg3}
\end{figure}
\begin{figure}
    \centering
    \includegraphics[]{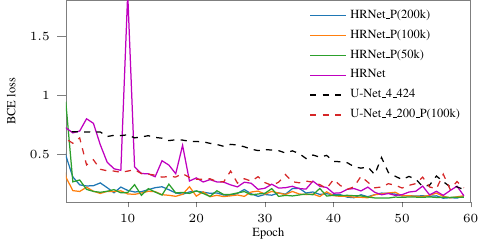}
    \vspace{-.4cm}
    \caption{Validation loss for HRNet models compared to U-Net.}
    \label{fig:seg4}
\end{figure}

In terms of model complexity, the observation is still consistent. As illustrated in Figure \ref{fig:seg3}, a randomly initialized U-Net\_4\_424 performs worse than the three different combinations of fine-tuned models.

\begin{figure}[!ht]
    \centering
    \includegraphics[width=.23\columnwidth]{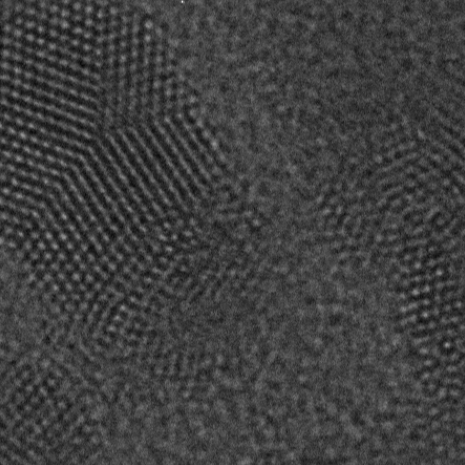}
    \includegraphics[width=.23\columnwidth]{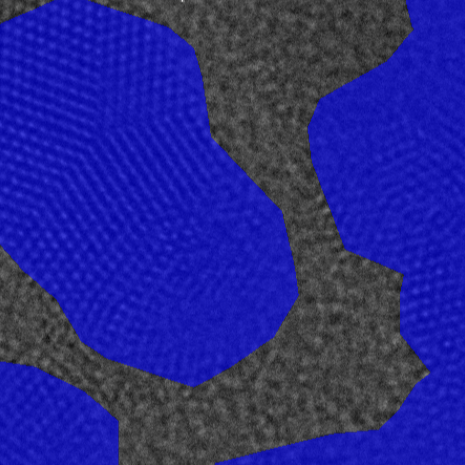}
    \includegraphics[width=.23\columnwidth]{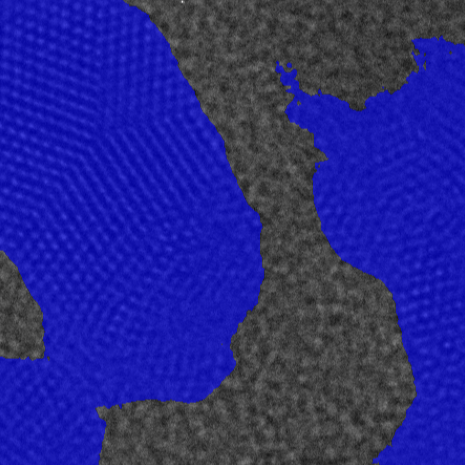}
    \includegraphics[width=.23\columnwidth]{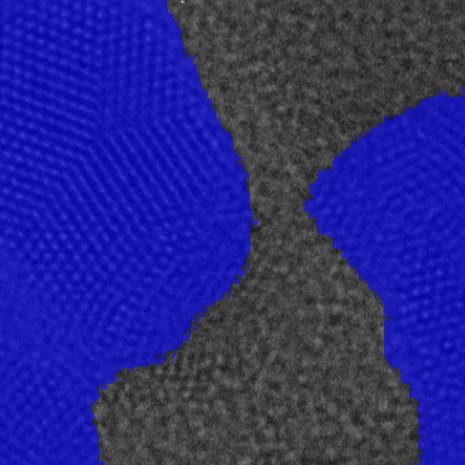}
    \caption{Left to write: original image, ground truth, prediction by fine-tuned HRNet and U-Net\_4\_0, respectively, both pretrained on \SI{100}{K} CEM500K images and fine-tuned.}
    \label{fig:ausegimages1}
\end{figure}

Conducting the same experiments with the HRNet model, we find that a fine-tuned U-Net model still outperforms the randomly initialized HRNet model, but all three fine-tuned HRNet models not only outperform the randomly initialized HRNet model but also score higher than the randomly initialized U-Net, as well as all fine-tuned U-Net models. The validation plots are shown in Figure \ref{fig:seg4}. Example predictions are illustrated in Figure~\ref{fig:ausegimages1} for the fine-tuned HRNet and fine-tuned U-Net\_4\_0, pretrained on \SI{100}{K} CEM500K images. Moreover, the dice scores on test set are calculated for all models after 5, 30 and 60 epochs. As shown in Table \ref{tab:smalltable}, the pretrained models converge faster than randomly initialized models and generally score higher, specially in the beginning epochs. The general trend of the results and observations for all other datasets is consistent with the above-reported ones.

\vspace{-0.2cm}
\begin{figure*}[!ht]
    \centering
    \includegraphics[width=0.32\textwidth]{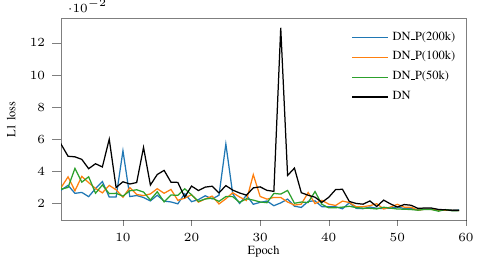}
    \includegraphics[width=0.32\textwidth]{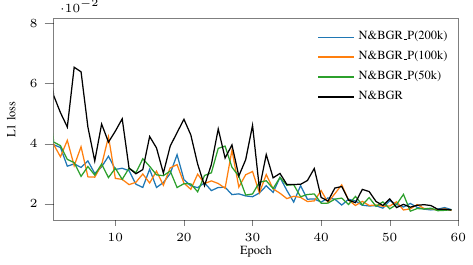}
    \includegraphics[width=0.32\textwidth]{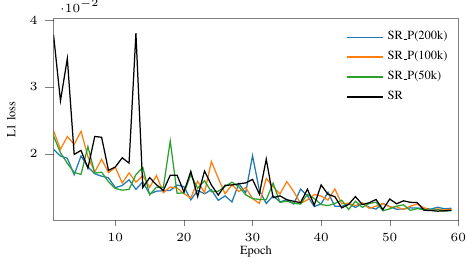}
    \caption{Validation $L_1$ loss for the downstream tasks on TEMImageNet dataset. Left to right:  denoising (DN), noise \& background removal (N\&BGR), and super-resolution (SR)}
    \label{fig:denois_l1}
\end{figure*}

\begin{table}[!ht]
\caption{Comparison of segmentation Dice scores for different training methods (randomly initialized weights (R) and pretrained (P) with GANs on CEM500K using \SI{50}{K}, \SI{100}{K}, \SI{200}{K} images) on the Au5nmV1 test dataset. The experiments were performed with UNets of different sizes (with two and four residual blocks) and receptive fields (two for each U-Net size) and HRNet.}
\centering
\begin{tabular}{lll|lll}
\multicolumn{3}{l|}{Epochs} & 5 & 30 & 60 \\ \hline
\multirow{4}{*}{\rotatebox[origin=c]{90}{HRNet}} & \multirow{4}{*}{} & R & 41.05 & 86.55 & 90.24 \\
 &  & P(50k) & 86.37 & 91 & 91.97 \\
 &  & P(100k) & 88.49 & 92.14 & 92.38 \\
 &  & P(200k) & 83.86 & 91.41 & 92.07 \\ \hline
\multirow{8}{*}{\rotatebox[origin=c]{90}{U-Net 4 blocks}} & \multirow{4}{*}{\rotatebox[origin=c]{90}{424}} & R & 7.8 & 86.03 & 89.95 \\
 &  & P(50k) & 72.78 & 89.81 & 91.16 \\
 &  & P(100k) & 71.8 & 89 & 89.72 \\
 &  & P(200k) & 75.46 & 87.34 & 90.21 \\
 & \multirow{4}{*}{\rotatebox[origin=c]{90}{200}} & R & 0.25 & 80.69 & 83.61 \\
 &  & P(50k) & 82.03 & 87.05 & 85.68 \\
 &  & P(100k) & 83 & 86.7 & 88.53 \\
 &  & P(200k) & 55.75 & 86.56 & 89 \\ \hline
\multirow{8}{*}{\rotatebox[origin=c]{90}{U-Net 2 blocks}} & \multirow{4}{*}{\rotatebox[origin=c]{90}{116}} & R & 0 & 80.81 & 82.49 \\
 &  & P(50k) & 61.99 & 82.3 & 84.72 \\
 &  & P(100k) & 51.97 & 80.16 & 83.47 \\
 &  & P(200k) & 67.87 & 80.48 & 83.18 \\
 & \multirow{4}{*}{\rotatebox[origin=c]{90}{44}} & R & 0 & 74.87 & 79.75 \\
 &  & P(50k) & 65.89 & 76.9 & 78.59 \\
 &  & P(100k) & 58.62 & 75.67 & 79.92 \\
 &  & P(200k) & 53.15 & 75.45 & 76.99
\end{tabular}%
\label{tab:smalltable}
\end{table}
\subsection{Experiments on TEMImageNet dataset} 

As the HRNet model performed better than all U-Net variations regardless of the fine-tuning, we conducted the experiments on the TEMImageNet dataset only with HRNet. On this dataset, we decided to exclusively employ HRNet for denoising, noise \& background removal, and super-resolution tasks. For each of these tasks, the HRNet model was trained with random initialization. The same experiments were conducted by fine-tuning the HRNet model pretrained on three separate subsets of the unlabeled CEM500K datasets in the previous step. 
During the experiments, an image size of 256x256 pixels and a batch size of 64 were used, and the training process continued for 60 epochs. The objective function comprised both $L_1$ and $L_2$ terms, and optimization was carried out using the Adam optimizer with a learning rate of \SI{2e-4}{}. The chosen data augmentation techniques, including noise variations, flipping, rotations, and random resizing, as usual had the goal of enhancing the model's ability to generalize and learn robust features. 
\begin{figure}
\centering
\begin{tikzpicture}
\node[anchor=south west,inner sep=0] (Bild) at (0,0)
{\includegraphics[width=.23\columnwidth, height=.23\columnwidth]{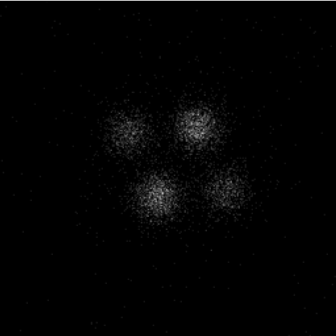}};
\begin{scope}[x=(Bild.south east),y=(Bild.north west)]
\node[color=white] at (0.1,0.9) {A};
\end{scope}
\end{tikzpicture}
\begin{tikzpicture}
\node[anchor=south west,inner sep=0] (Bild) at (0,0)
{\includegraphics[width=.23\columnwidth, height=.23\columnwidth]
{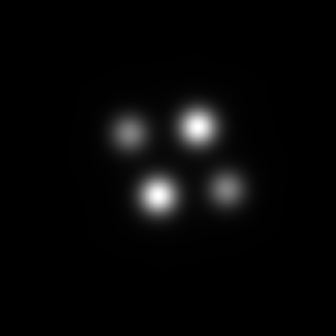}};
\begin{scope}[x=(Bild.south east),y=(Bild.north west)]
\node[color=white] at (0.1,0.9) {B};
\end{scope}
\end{tikzpicture}
\begin{tikzpicture}
\node[anchor=south west,inner sep=0] (Bild) at (0,0)
{\includegraphics[width=.23\columnwidth, height=.23\columnwidth]{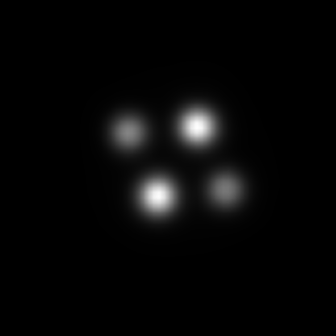}};
\begin{scope}[x=(Bild.south east),y=(Bild.north west)]
\node[color=white] at (0.1,0.9) {C};
\end{scope}
\end{tikzpicture}\\
\begin{tikzpicture}
\node[anchor=south west,inner sep=0] (Bild) at (0,0)
{\includegraphics[width=.23\columnwidth, height=.23\columnwidth]{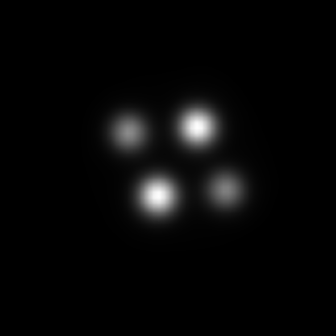}};
\begin{scope}[x=(Bild.south east),y=(Bild.north west)]
\node[color=white] at (0.1,0.9) {D};
\end{scope}
\end{tikzpicture}
\begin{tikzpicture}
\node[anchor=south west,inner sep=0] (Bild) at (0,0)
{\includegraphics[width=.23\columnwidth, height=.23\columnwidth]
{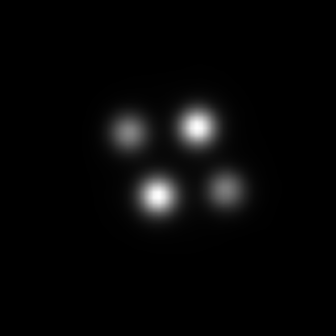}};
\begin{scope}[x=(Bild.south east),y=(Bild.north west)]
\node[color=white] at (0.1,0.9) {E};
\end{scope}
\end{tikzpicture}
\begin{tikzpicture}
\node[anchor=south west,inner sep=0] (Bild) at (0,0)
{\includegraphics[width=.23\columnwidth, height=.23\columnwidth]{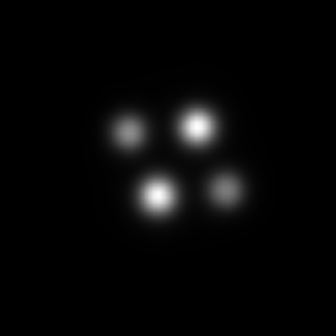}};
\begin{scope}[x=(Bild.south east),y=(Bild.north west)]
\node[color=white] at (0.1,0.9) {F};
\end{scope}
\end{tikzpicture}\\
\caption{Results for denoising on TEMImageNet. Input (A), ground truth (B), prediction by randomly initialized model (C), and predictions (D, E, F) by fine-tuned models pretrained on $50$K, $100$K, and $200$K images.}
\label{fig:nonoiseprediction}
\end{figure}
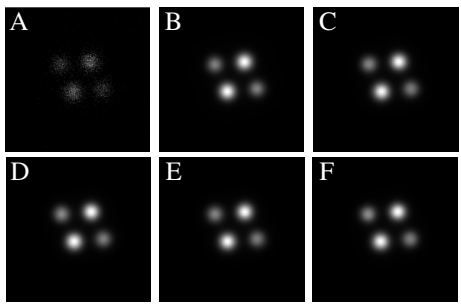

\begin{figure}[!ht]
    \centering
    \begin{tikzpicture}
    \node[anchor=south west,inner sep=0] (Bild) at (0,0)
    {
    \includegraphics[width=.23\columnwidth, height=.23\columnwidth]{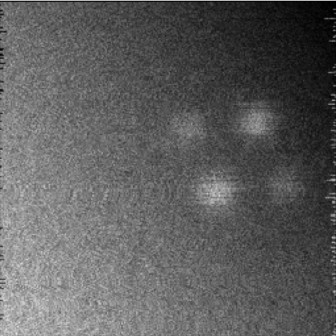}};
    \begin{scope}[x=(Bild.south east),y=(Bild.north west)]
    \node[color=white] at (0.1,0.9) {A};
    \end{scope}
    \end{tikzpicture}
    \begin{tikzpicture}
    \node[anchor=south west,inner sep=0] (Bild) at (0,0)
    {
    \includegraphics[width=.23\columnwidth, height=.23\columnwidth]{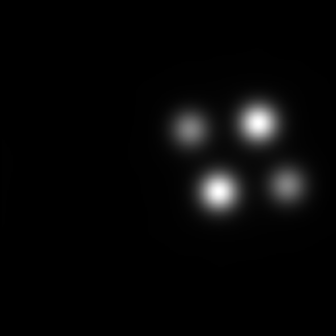}};
    \begin{scope}[x=(Bild.south east),y=(Bild.north west)]
    \node[color=white] at (0.1,0.9) {B};
    \end{scope}
    \end{tikzpicture}
    \begin{tikzpicture}
    \node[anchor=south west,inner sep=0] (Bild) at (0,0)
    {
    \includegraphics[width=.23\columnwidth, height=.23\columnwidth]{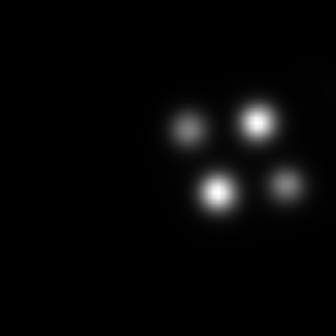}};
    \begin{scope}[x=(Bild.south east),y=(Bild.north west)]
    \node[color=white] at (0.1,0.9) {C};
    \end{scope}
    \end{tikzpicture}\\
    \begin{tikzpicture}
    \node[anchor=south west,inner sep=0] (Bild) at (0,0)
    {
    \includegraphics[width=.23\columnwidth, height=.23\columnwidth]{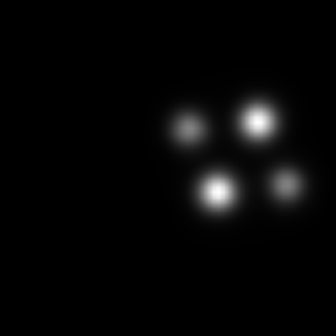}};
    \begin{scope}[x=(Bild.south east),y=(Bild.north west)]
    \node[color=white] at (0.1,0.9) {D};
    \end{scope}
    \end{tikzpicture}
    \begin{tikzpicture}
    \node[anchor=south west,inner sep=0] (Bild) at (0,0)
    {
    \includegraphics[width=.23\columnwidth, height=.23\columnwidth]{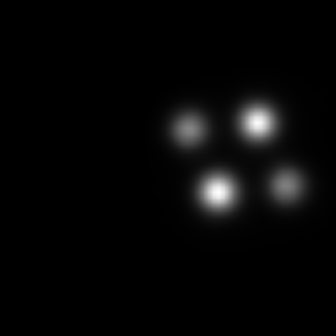}};
    \begin{scope}[x=(Bild.south east),y=(Bild.north west)]
    \node[color=white] at (0.1,0.9) {E};
    \end{scope}
    \end{tikzpicture}
    \begin{tikzpicture}
    \node[anchor=south west,inner sep=0] (Bild) at (0,0)
    {
    \includegraphics[width=.23\columnwidth, height=.23\columnwidth]{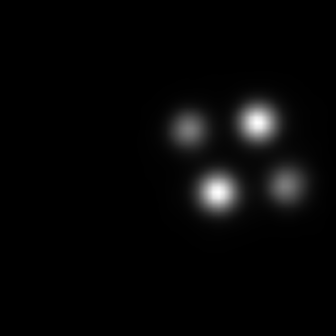}};
    \begin{scope}[x=(Bild.south east),y=(Bild.north west)]
    \node[color=white] at (0.1,0.9) {F};
    \end{scope}
    \end{tikzpicture}\\
    \caption{Results for noise \& background removal on TEMImageNet. 
    (A)-(F) in analogy to Figure~\ref{fig:nonoiseprediction}.
    }
    \label{fig:nobackgroundprediction}
\end{figure}
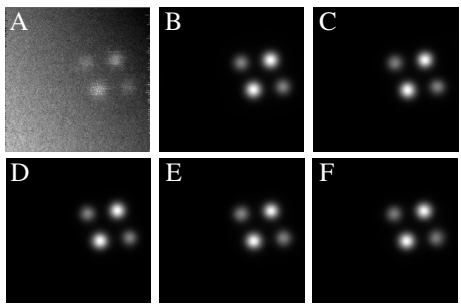

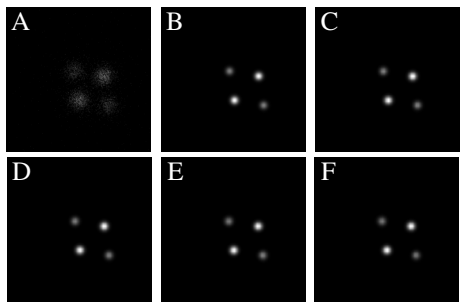
\begin{figure}[!ht]
    \centering
        \begin{tikzpicture}
    \node[anchor=south west,inner sep=0] (Bild) at (0,0)
    {
    \includegraphics[width=.23\columnwidth, height=.23\columnwidth]{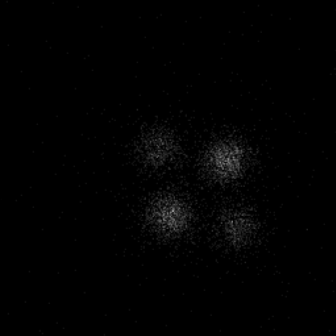}};
    \begin{scope}[x=(Bild.south east),y=(Bild.north west)]
    \node[color=white] at (0.1,0.9) {A};
    \end{scope}
    \end{tikzpicture}
        \begin{tikzpicture}
    \node[anchor=south west,inner sep=0] (Bild) at (0,0)
    {
    \includegraphics[width=.23\columnwidth, height=.23\columnwidth]{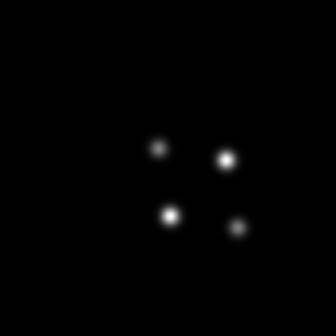}};
    \begin{scope}[x=(Bild.south east),y=(Bild.north west)]
    \node[color=white] at (0.1,0.9) {B};
    \end{scope}
    \end{tikzpicture}
        \begin{tikzpicture}
    \node[anchor=south west,inner sep=0] (Bild) at (0,0)
    {
    \includegraphics[width=.23\columnwidth, height=.23\columnwidth]{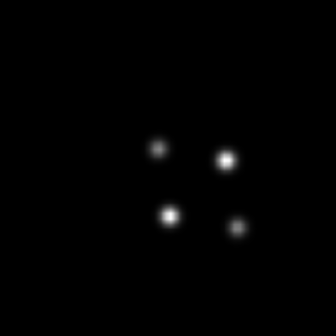}};
    \begin{scope}[x=(Bild.south east),y=(Bild.north west)]
    \node[color=white] at (0.1,0.9) {C};
    \end{scope}
    \end{tikzpicture}\\
        \begin{tikzpicture}
    \node[anchor=south west,inner sep=0] (Bild) at (0,0)
    {
    \includegraphics[width=.23\columnwidth, height=.23\columnwidth]{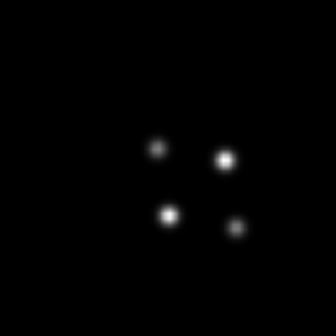}};
    \begin{scope}[x=(Bild.south east),y=(Bild.north west)]
    \node[color=white] at (0.1,0.9) {D};
    \end{scope}
    \end{tikzpicture}
        \begin{tikzpicture}
    \node[anchor=south west,inner sep=0] (Bild) at (0,0)
    {
    \includegraphics[width=.23\columnwidth, height=.23\columnwidth]{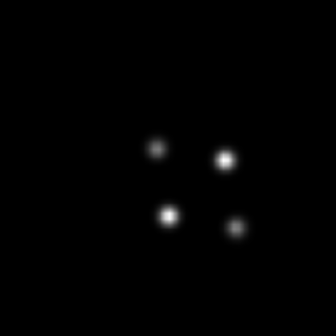}};
    \begin{scope}[x=(Bild.south east),y=(Bild.north west)]
    \node[color=white] at (0.1,0.9) {E};
    \end{scope}
    \end{tikzpicture}
        \begin{tikzpicture}
    \node[anchor=south west,inner sep=0] (Bild) at (0,0)
    {
    \includegraphics[width=.23\columnwidth, height=.23\columnwidth]{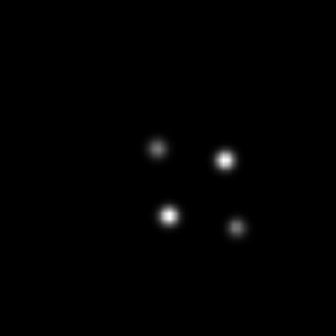}
    };
    \begin{scope}[x=(Bild.south east),y=(Bild.north west)]
    \node[color=white] at (0.1,0.9) {F};
    \end{scope}
    \end{tikzpicture}\\
    \caption{Results for super-resolution on TEMImageNet. 
    (A)-(F) in analogy to Figure~\ref{fig:nonoiseprediction}.
    }
    \label{fig:superresolutionprediction}
\end{figure}

The plots for validation loss in all three tasks are shown in Figure ~\ref{fig:denois_l1}. We observe that in all three cases, the validation losses for the fine-tuned models are lower than those for the randomly initialized models. 
During the initial training phase, the validation loss for the model fine-tuned with smaller datasets is larger than that for the models trained with more data. However, already after approximately 10 epochs, this difference vanishes. Additionally, the validation $L_1$ loss of all randomly initialized models exhibits severe fluctuations, while the fine-tuned models behave much more robustly. Some example predictions for denoising, noise \& background removal, and super-resolution are shown in Figures~\ref{fig:nonoiseprediction}, \ref{fig:nobackgroundprediction}, and \ref{fig:superresolutionprediction}, respectively. 
Even though all predictions are visually very good and it is difficult to observe any differences, the quantitative results as confirmed by the plots in Figure \ref{fig:denois_l1} show that pretraining leads to a better and more stable performance. Similarly, the $L_1$ metric on test set are calculated for all models after 5, 30 and 60 epochs. As shown in Table \ref{tab:smalltabletem}, the pretrained models converge faster than randomly intialized models and generally score better, specially in the beginning epochs.

\begin{table}[]
\caption{Comparison of the $L_1$ metric for different training methods (randomly initialized weights (R) and pretrained (P) with GANs on CEM500K using \SI{50}{K}, \SI{100}{K}, \SI{200}{K} images) on each of the downstream tasks: Super-resolution (SR), Noise \& Background Removal (N\&BGR) and Denoising (DN). The experiments were performed with HRNet.}
\begin{tabular}{ll|lll}
\multicolumn{2}{l|}{Epochs} & 5 & 30 & 60 \\ \hline
\multirow{4}{*}{SR} & R & 0.01972 & 0.01524 & 0.01122 \\
 & P(50k) & 0.01688 & 0.01359 & 0.0112 \\
 & P(100k) & 0.02105 & 0.01465 & 0.01142 \\
 & P(200k) & 0.01639 & 0.01371 & 0.01156 \\ \hline
\multirow{4}{*}{N\&BGR} & R & 0.06391 & 0.03438 & 0.0176 \\
 & P(50k) & 0.03332 & 0.02257 & 0.01776 \\
 & P(100k) & 0.03268 & 0.02913 & 0.01766 \\
 & P(200k) & 0.03323 & 0.0224 & 0.01792 \\ \hline
\multirow{4}{*}{DN} & R & 0.04732 & 0.02996 & 0.01541 \\
 & P(50k) & 0.03287 & 0.02169 & 0.01542 \\
 & P(100k) & 0.03693 & 0.03708 & 0.01555 \\
 & P(200k) & 0.02657 & 0.01907 & 0.01586
\end{tabular}%
\label{tab:smalltabletem}
\end{table}

\section{Conclusion}
\label{sec:conclusion}

In this paper, we explored the impact of pretraining on various computer vision tasks. Through self-supervised training, GANs, and a pretraining strategy involving unlabeled data followed by fine-tuning on labeled data, our investigation showcased significant advancements in the capabilities of computer vision models, specifically in the context of EM.
Self-supervised training enabled the models to extract representations from unlabeled EM data, addressing challenges associated with the scarcity and labor intensity of labeled datasets in this domain. Furthermore, the integration of GANs in generative pretraining proved beneficial for improving model generalization.

Pretraining on unlabeled data, followed by fine-tuning on labeled data, enhanced performance and accelerated convergence in several downstream tasks, including segmentation, denoising, and super-resolution. An important outcome of our work for such tasks is that for obtaining a higher predictive accuracy, the model complexity might not be the only or most important factor. The CEM500K dataset, containing SEM images, was used for pretraining and improved performance in TEM image-based downstream tasks, despite the differences between SEM and TEM images. Future research could explore pretraining on TEM images for closer domain relevance. Additionally, while self-supervised learning with GANs enhanced the performance in this study, their training complexities and risk of mode collapse suggest exploring alternative self-supervised methods like contrastive pretraining for potentially better outcomes in the future.

{
    \small
    \bibliographystyle{ieeenat_fullname}
    \bibliography{main}

\begin{thebibliography}{56}
\providecommand{\natexlab}[1]{#1}
\providecommand{\url}[1]{\texttt{#1}}
\expandafter\ifx\csname urlstyle\endcsname\relax
  \providecommand{\doi}[1]{doi: #1}\else
  \providecommand{\doi}{doi: \begingroup \urlstyle{rm}\Url}\fi

\bibitem[Aguiar et~al.(2019)Aguiar, Gong, Unocic, Taşdizen, and
  Miller]{10.1126/sciadv.aaw1949}
J.~A. Aguiar, M.~L. Gong, R.~R. Unocic, T. Taşdizen, and B. Miller.
\newblock Decoding crystallography from high-resolution electron imaging and
  diffraction datasets with deep learning.
\newblock \emph{Science Advances}, 5, 2019.

\bibitem[Alosaimi et~al.(2023)Alosaimi, Alhichri, Bazi, Youssef, and
  Alajlan]{10.1038/s41598-022-27313-5}
N. Alosaimi, H. Alhichri, Y. Bazi, B.~B. Youssef, and N. Alajlan.
\newblock Self-supervised learning for remote sensing scene classification
  under the few shot scenario.
\newblock \emph{Scientific Reports}, 13, 2023.

\bibitem[Angermueller et~al.(2016)Angermueller, Pärnamaa, and
  Parts]{10.15252/msb.20156651}
C. Angermueller, T. Pärnamaa, and L. Parts.
\newblock Deep learning for computational biology.
\newblock \emph{Molecular Systems Biology}, 12, 2016.

\bibitem[Aswath et~al.(2023)Aswath, Alsahaf, Giepmans, and
  Azzopardi]{ASWATH2023102920}
Anusha Aswath, Ahmad Alsahaf, Ben~NG Giepmans, and George Azzopardi.
\newblock Segmentation in large-scale cellular electron microscopy with deep
  learning: A literature survey.
\newblock \emph{Medical image analysis}, page 102920, 2023.

\bibitem[Caron et~al.(2021)Caron, Touvron, Misra, J\'egou, Mairal, Bojanowski,
  and Joulin]{Caron_2021_ICCV}
Mathilde Caron, Hugo Touvron, Ishan Misra, Herv\'e J\'egou, Julien Mairal,
  Piotr Bojanowski, and Armand Joulin.
\newblock Emerging properties in self-supervised vision transformers.
\newblock In \emph{Proceedings of the IEEE/CVF International Conference on
  Computer Vision (ICCV)}, pages 9650--9660, 2021.

\bibitem[Chen et~al.(2019)Chen, Zhai, Ritter, Lucic, and Houlsby]{chen2019self}
Ting Chen, Xiaohua Zhai, Marvin Ritter, Mario Lucic, and Neil Houlsby.
\newblock Self-supervised {GAN}s via auxiliary rotation loss.
\newblock In \emph{Proceedings of the IEEE/CVF conference on computer vision
  and pattern recognition}, pages 12154--12163, 2019.

\bibitem[Chen et~al.(2020{\natexlab{a}})Chen, Kornblith, Norouzi, and
  Hinton]{chen2020simple}
Ting Chen, Simon Kornblith, Mohammad Norouzi, and Geoffrey Hinton.
\newblock A simple framework for contrastive learning of visual
  representations.
\newblock In \emph{International conference on machine learning}, pages
  1597--1607. PMLR, 2020{\natexlab{a}}.

\bibitem[Chen et~al.(2020{\natexlab{b}})Chen, Fan, Girshick, and He]{Chen2020}
Xinlei Chen, Haoqi Fan, Ross~B. Girshick, and Kaiming He.
\newblock Improved baselines with momentum contrastive learning.
\newblock \emph{ArXiv}, abs/2003.04297, 2020{\natexlab{b}}.

\bibitem[Conrad and Narayan(2021)]{Conrad2021}
Ryan Conrad and Kedar Narayan.
\newblock Cem500k, a large-scale heterogeneous unlabeled cellular electron
  microscopy image dataset for deep learning.
\newblock \emph{eLife}, 10, 2021.

\bibitem[Ding et~al.(2020)Ding, He, and Wan]{10.48550/arxiv.2001.01986}
Ke Ding, Xuanji He, and Guanglu Wan.
\newblock Learning speaker embedding with momentum contrast.
\newblock \emph{ArXiv}, abs/2001.01986, 2020.

\bibitem[Ezzat and Bedewy(2020)]{10.1021/acs.jpcc.0c07765}
A.~A. Ezzat and M. Bedewy.
\newblock Machine learning for revealing spatial dependence among
  nanoparticles: understanding catalyst film dewetting via gibbs point process
  models.
\newblock \emph{The Journal of Physical Chemistry C}, 124:\penalty0
  27479--27494, 2020.

\bibitem[Goodfellow et~al.(2014)Goodfellow, Pouget-Abadie, Mirza, Xu,
  Warde-Farley, Ozair, Courville, and Bengio]{goodfellow2014generative}
Ian Goodfellow, Jean Pouget-Abadie, Mehdi Mirza, Bing Xu, David Warde-Farley,
  Sherjil Ozair, Aaron Courville, and Yoshua Bengio.
\newblock Generative adversarial nets.
\newblock \emph{Advances in neural information processing systems}, 27, 2014.

\bibitem[Govind et~al.(2024)Govind, Oliveros, Dlouhy, Legros, and
  Sandfeld]{Govind_Oliveros_Dlouhy_Legros_Sandfeld_2024}
Kishan Govind, Daniela Oliveros, Antonin Dlouhy, Marc Legros, and Stefan
  Sandfeld.
\newblock Deep learning of crystalline defects from {TEM} images: a solution
  for the problem of ‘never enough training data’.
\newblock \emph{Machine Learning: Science and Technology}, 5\penalty0
  (1):\penalty0 015006, 2024.

\bibitem[Groschner et~al.(2021)Groschner, Choi, and Scott]{Groschner2021}
Catherine~K. Groschner, Christina Choi, and Mary~C. Scott.
\newblock Machine learning pipeline for segmentation and defect identification
  from high-resolution transmission electron microscopy data.
\newblock \emph{Microscopy and Microanalysis}, 27\penalty0 (3):\penalty0
  549–556, 2021.

\bibitem[Guo et~al.(2021)Guo, Xia, and Luo]{guo2021self}
Dongen Guo, Ying Xia, and Xiaobo Luo.
\newblock Self-supervised {GANs} with similarity loss for remote sensing image
  scene classification.
\newblock \emph{IEEE Journal of Selected Topics in Applied Earth Observations
  and Remote Sensing}, 14:\penalty0 2508--2521, 2021.

\bibitem[He et~al.(2020)He, Fan, Wu, Xie, and Girshick]{he2020momentum}
Kaiming He, Haoqi Fan, Yuxin Wu, Saining Xie, and Ross Girshick.
\newblock Momentum contrast for unsupervised visual representation learning.
\newblock In \emph{Proceedings of the IEEE/CVF conference on computer vision
  and pattern recognition}, pages 9729--9738, 2020.

\bibitem[Horwath et~al.(2020)Horwath, Zakharov, Mégret, and
  Stach]{Horwath2020}
James~P. Horwath, Dmitri~N. Zakharov, Rémi Mégret, and Eric~A. Stach.
\newblock Understanding important features of deep learning models for
  segmentation of high-resolution transmission electron microscopy images.
\newblock \emph{npj Computational Materials}, 6\penalty0 (1), 2020.

\bibitem[Isola et~al.(2017)Isola, Zhu, Zhou, and Efros]{Isola2017}
Phillip Isola, Jun-Yan Zhu, Tinghui Zhou, and Alexei~A. Efros.
\newblock {Image-to-Image} translation with conditional adversarial networks.
\newblock In \emph{2017 IEEE Conference on Computer Vision and Pattern
  Recognition (CVPR)}. IEEE, 2017.

\bibitem[Iwata et~al.(2022)Iwata, Hayashi, Hasegawa, Terayama, and
  Okuno]{Iwata2022}
Hiroaki Iwata, Yoshihiro Hayashi, Aki Hasegawa, Kei Terayama, and Yasushi
  Okuno.
\newblock Classification of scanning electron microscope images of
  pharmaceutical excipients using deep convolutional neural networks with
  transfer learning.
\newblock \emph{International Journal of Pharmaceutics: X}, 4:\penalty0 100135,
  2022.

\bibitem[Kazimi and Sester(2023)]{kazimi2023self}
Bashir Kazimi and Monika Sester.
\newblock Self-supervised learning for semantic segmentation of archaeological
  monuments in {DTMs}.
\newblock \emph{Journal of computer applications in archaeology 6 (2023), Nr.
  1}, 6\penalty0 (1):\penalty0 155--173, 2023.

\bibitem[Ke(1971)]{10.1016/s0074-7696(08)60048-0}
C. Ke.
\newblock Applications of scanning electron microscopy in biology.
\newblock \emph{International Review of Cytology}, pages 183--255, 1971.

\bibitem[Khadangi et~al.(2021)Khadangi, Boudier, and Rajagopal]{Khadangi2021}
Afshin Khadangi, Thomas Boudier, and Vijay Rajagopal.
\newblock {EM-net}: Deep learning for electron microscopy image segmentation.
\newblock In \emph{2020 25th International Conference on Pattern Recognition
  (ICPR)}. IEEE, 2021.

\bibitem[Lee et~al.(2020)Lee, Yoon, Lee, Kim, Chang, Yun, Ro, Lee, and
  Lee]{10.1021/acsnano.0c06809}
B. Lee, S. Yoon, J.~W. Lee, Y. Kim, J. Chang, J. Yun, J.~C. Ro, J. Lee, and
  J.~H. Lee.
\newblock Statistical characterization of the morphologies of nanoparticles
  through machine learning based electron microscopy image analysis.
\newblock \emph{ACS Nano}, 14:\penalty0 17125--17133, 2020.

\bibitem[Leng(2013)]{leng2013materials}
Yang Leng.
\newblock \emph{Materials characterization: introduction to microscopic and
  spectroscopic methods}.
\newblock John Wiley \& Sons, 2013.

\bibitem[Li et~al.(2022)Li, Togo, Ogawa, and
  Haseyama]{10.48550/arxiv.2211.00313}
Guang Li, Ren Togo, Takahiro Ogawa, and Miki Haseyama.
\newblock {RGMIM}: Region-guided masked image modeling for {COVID-19}
  detection.
\newblock \emph{arXiv e-prints}, pages arXiv--2211, 2022.

\bibitem[Lin et~al.(2021)Lin, Zhang, Wang, Yang, and Xin]{Lin2021}
Ruoqian Lin, Rui Zhang, Chunyang Wang, Xiao-Qing Yang, and Huolin~L. Xin.
\newblock {TEMImageNet} training library and atomsegnet deep-learning models
  for high-precision atom segmentation, localization, denoising, and deblurring
  of atomic-resolution images.
\newblock \emph{Scientific Reports}, 11\penalty0 (1), 2021.

\bibitem[Lupan et~al.(2014)Lupan, Creţu, Deng, Gedamu, Paulowicz, Kaps,
  Mishra, Polonskyi, Zamponi, Kienle, Trofim, Tiginyanu, and
  Adelung]{10.1021/jp5038415}
O. Lupan, V. Creţu, M. Deng, D. Gedamu, I. Paulowicz, S. Kaps, Y.~K. Mishra,
  O. Polonskyi, C. Zamponi, L. Kienle, V. Trofim, I.~M. Tiginyanu, and R.
  Adelung.
\newblock Versatile growth of freestanding orthorhombic $\alpha$-molybdenum
  trioxide nano- and microstructures by rapid thermal processing for gas
  nanosensors.
\newblock \emph{The Journal of Physical Chemistry C}, 118:\penalty0
  15068--15078, 2014.

\bibitem[Mao et~al.(2017)Mao, Li, Xie, Lau, Wang, and
  Paul~Smolley]{mao2017least}
Xudong Mao, Qing Li, Haoran Xie, Raymond~YK Lau, Zhen Wang, and Stephen
  Paul~Smolley.
\newblock Least squares generative adversarial networks.
\newblock In \emph{Proceedings of the IEEE international conference on computer
  vision}, pages 2794--2802, 2017.

\bibitem[Marmanis et~al.(2015)Marmanis, Datcu, Esch, and
  Stilla]{marmanis2015deep}
Dimitrios Marmanis, Mihai Datcu, Thomas Esch, and Uwe Stilla.
\newblock Deep learning earth observation classification using imagenet
  pretrained networks.
\newblock \emph{IEEE Geoscience and Remote Sensing Letters}, 13\penalty0
  (1):\penalty0 105--109, 2015.

\bibitem[Mirza and Osindero(2014)]{10.48550/arxiv.1411.1784}
Mehdi Mirza and Simon Osindero.
\newblock Conditional generative adversarial nets.
\newblock \emph{arXiv preprint arXiv:1411.1784}, 2014.

\bibitem[Oh et~al.(2009)Oh, Legros, Kiener, and
  Dehm]{Oh_Legros_Kiener_Dehm_2009}
Sang~Ho Oh, Marc Legros, Daniel Kiener, and Gerhard Dehm.
\newblock In situ observation of dislocation nucleation and escape in a
  submicrometre aluminium single crystal.
\newblock \emph{Nature materials}, 8\penalty0 (2):\penalty0 95--100, 2009.

\bibitem[Panova et~al.(2016)Panova, Chen, Bustillo, Ophus, Bhatt, Balsara, and
  Minor]{10.1016/j.micron.2016.05.008}
O. Panova, X.~C. Chen, K.~C. Bustillo, C. Ophus, M.~P. Bhatt, N.~P. Balsara,
  and A.~M. Minor.
\newblock Orientation mapping of semicrystalline polymers using scanning
  electron nanobeam diffraction.
\newblock \emph{Micron}, 88:\penalty0 30--36, 2016.

\bibitem[Patel et~al.(2021)Patel, Kumari, Singh, and
  Krishnamurthy]{patel2021lt}
Parth Patel, Nupur Kumari, Mayank Singh, and Balaji Krishnamurthy.
\newblock {LT-GAN}: Self-supervised {GAN} with latent transformation detection.
\newblock In \emph{Proceedings of the IEEE/CVF winter conference on
  applications of computer vision}, pages 3189--3198, 2021.

\bibitem[Pathak et~al.(2016)Pathak, Krahenbuhl, Donahue, Darrell, and
  Efros]{pathak2016context}
Deepak Pathak, Philipp Krahenbuhl, Jeff Donahue, Trevor Darrell, and Alexei~A
  Efros.
\newblock Context encoders: {Feature} learning by inpainting.
\newblock In \emph{Proceedings of the IEEE conference on computer vision and
  pattern recognition}, pages 2536--2544, 2016.

\bibitem[Paul et~al.(2016)Paul, Hawkins, Balagurunathan, Schabath, Gillies,
  Hall, and Goldgof]{10.18383/j.tom.2016.00211}
R. Paul, S. Hawkins, Y. Balagurunathan, M.~B. Schabath, R.~J. Gillies, L.~O.
  Hall, and D.~B. Goldgof.
\newblock Deep feature transfer learning in combination with traditional
  features predicts survival among patients with lung adenocarcinoma.
\newblock \emph{Tomography}, 2:\penalty0 388--395, 2016.

\bibitem[Ridnik et~al.(2021)Ridnik, Ben-Baruch, Noy, and
  Zelnik-Manor]{ridnik2021imagenet}
Tal Ridnik, Emanuel Ben-Baruch, Asaf Noy, and Lihi Zelnik-Manor.
\newblock {ImageNet-21K} pretraining for the masses.
\newblock In \emph{Thirty-fifth Conference on Neural Information Processing
  Systems Datasets and Benchmarks Track (Round 1)}, 2021.

\bibitem[Ronneberger et~al.(2015)Ronneberger, Fischer, and
  Brox]{ronneberger2015u}
Olaf Ronneberger, Philipp Fischer, and Thomas Brox.
\newblock {U-Net}: Convolutional networks for biomedical image segmentation.
\newblock In \emph{Medical Image Computing and Computer-Assisted
  Intervention--MICCAI 2015: 18th International Conference, Munich, Germany,
  October 5-9, 2015, Proceedings, Part III 18}, pages 234--241. Springer, 2015.

\bibitem[Ruzaeva et~al.(2023)Ruzaeva, Govind, Legros, and
  Sandfeld]{Ruzaeva_Govind_Legros_Sandfeld_2023}
Karina Ruzaeva, Kishan Govind, Marc Legros, and Stefan Sandfeld.
\newblock Instance segmentation of dislocations in {TEM} images.
\newblock In \emph{2023 IEEE 23rd International Conference on Nanotechnology
  (NANO)}, pages 1--6. IEEE, 2023.

\bibitem[Sadre et~al.(2021)Sadre, Ophus, Butko, and Weber]{sadre2021deep}
Robbie Sadre, Colin Ophus, Anastasiia Butko, and Gunther~H Weber.
\newblock Deep learning segmentation of complex features in atomic-resolution
  phase-contrast transmission electron microscopy images.
\newblock \emph{Microscopy and microanalysis}, 27\penalty0 (4):\penalty0
  804--814, 2021.

\bibitem[Salas et~al.(2017)Salas, Gall, Fiche, Valeri, Ke, Bron, Bellot, and
  Nollmann]{10.1073/pnas.1704908114}
D. Salas, A.~L. Gall, J. Fiche, A. Valeri, Y. Ke, P. Bron, G. Bellot, and M.
  Nollmann.
\newblock Angular reconstitution-based {3D} reconstructions of nanomolecular
  structures from superresolution light-microscopy images.
\newblock \emph{Proceedings of the National Academy of Sciences}, 114:\penalty0
  9273--9278, 2017.

\bibitem[Sowrirajan et~al.(2020)Sowrirajan, Yang, Ng, and
  Rajpurkar]{10.48550/arxiv.2010.05352}
H. Sowrirajan, J. Yang, A.~Y. Ng, and P. Rajpurkar.
\newblock {MoCo-CXR}: {MoCo} pretraining improves representation and
  transferability of chest {X-ray} models.
\newblock 2020.

\bibitem[Srinivas et~al.(2022)Srinivas, S., Zakariah, Alotaibi, Shaukat,
  Partibane, and Halifa]{10.1155/2022/3264367}
C. Srinivas, N.~P.~K. S., M. Zakariah, Y.~A. Alotaibi, K. Shaukat, B.
  Partibane, and A. Halifa.
\newblock Deep transfer learning approaches in performance analysis of brain
  tumor classification using {MRI} images.
\newblock \emph{Journal of Healthcare Engineering}, 2022:\penalty0 1--17, 2022.

\bibitem[Steinberger et~al.(2023)Steinberger, Issa, Strobl, Imrich, Kiener, and
  Sandfeld]{Steinberger_Issa_Strobl_Imrich_Kiener_Sandfeld_2023}
Dominik Steinberger, Inas Issa, Rachel Strobl, Peter~J Imrich, Daniel Kiener,
  and Stefan Sandfeld.
\newblock Data-mining of in-situ {TEM} experiments: Towards understanding
  nanoscale fracture.
\newblock \emph{Computational materials science}, 216:\penalty0 111830, 2023.

\bibitem[Stuckner et~al.(2022)Stuckner, Harder, and Smith]{Stuckner2022}
Joshua Stuckner, Bryan Harder, and Timothy~M. Smith.
\newblock Microstructure segmentation with deep learning encoders pre-trained
  on a large microscopy dataset.
\newblock \emph{npj Computational Materials}, 8\penalty0 (1), 2022.

\bibitem[Sun et~al.(2019)Sun, Xiao, Liu, and Wang]{SunXLW19}
Ke Sun, Bin Xiao, Dong Liu, and Jingdong Wang.
\newblock Deep high-resolution representation learning for human pose
  estimation.
\newblock In \emph{Proceedings of the IEEE/CVF Conference on Computer Vision
  and Pattern Recognition}, pages 5693--5703, 2019.

\bibitem[Sytwu et~al.(2022)Sytwu, Groschner, and Scott]{Sytwu2022}
Katherine Sytwu, Catherine Groschner, and Mary~C Scott.
\newblock Understanding the influence of receptive field and network complexity
  in neural network-guided {TEM} image analysis.
\newblock \emph{Microscopy and Microanalysis}, 28\penalty0 (6):\penalty0
  1896–1904, 2022.

\bibitem[Treneska et~al.(2022)Treneska, Zdravevski, Pires, Lameski, and
  Gievska]{treneska2022gan}
Sandra Treneska, Eftim Zdravevski, Ivan~Miguel Pires, Petre Lameski, and Sonja
  Gievska.
\newblock {GAN}-based image colorization for self-supervised visual feature
  learning.
\newblock \emph{Sensors}, 22\penalty0 (4):\penalty0 1599, 2022.

\bibitem[Valverde et~al.(2021)Valverde, Imani, Abdollahzadeh, De~Feo, Prakash,
  Ciszek, and Tohka]{10.48550/arxiv.2102.01530}
Juan~Miguel Valverde, Vandad Imani, Ali Abdollahzadeh, Riccardo De~Feo,
  Mithilesh Prakash, Robert Ciszek, and Jussi Tohka.
\newblock Transfer learning in magnetic resonance brain imaging: {A} systematic
  review.
\newblock \emph{Journal of imaging}, 7\penalty0 (4):\penalty0 66, 2021.

\bibitem[Vincent et~al.(2021)Vincent, Manzorro, Mohan, Tang, Sheth, Simoncelli,
  Matteson, Fernandez‐Granda, and Crozier]{10.1017/s1431927621012678}
J. Vincent, R. Manzorro, S. Mohan, B. Tang, D.~Y. Sheth, E.~P. Simoncelli,
  D.~S. Matteson, C. Fernandez‐Granda, and P.~A. Crozier.
\newblock Developing and evaluating deep neural network-based denoising for
  nanoparticle {TEM} images with ultra-low signal-to-noise.
\newblock \emph{Microscopy and Microanalysis}, 27:\penalty0 1431--1447, 2021.

\bibitem[Wang et~al.(2021)Wang, Sun, Cheng, Jiang, Deng, Zhao, Liu, Mu, Tan,
  Wang, Liu, and Xiao]{wang2021}
J. Wang, K. Sun, T. Cheng, B. Jiang, C. Deng, Y. Zhao, D. Liu, Y. Mu, M. Tan,
  X. Wang, W. Liu, and B. Xiao.
\newblock Deep high-resolution representation learning for visual recognition.
\newblock \emph{IEEE Transactions on Pattern Analysis \& Machine Intelligence},
  43\penalty0 (10):\penalty0 3349--3364, 2021.

\bibitem[Xie et~al.(2022)Xie, Zhang, Cao, Lin, Bao, Yao, Dai, and
  Hu]{xie2022simmim}
Zhenda Xie, Zheng Zhang, Yue Cao, Yutong Lin, Jianmin Bao, Zhuliang Yao, Qi
  Dai, and Han Hu.
\newblock {SimMIM}: A simple framework for masked image modeling.
\newblock In \emph{Proceedings of the IEEE/CVF Conference on Computer Vision
  and Pattern Recognition}, pages 9653--9663, 2022.

\bibitem[Xue et~al.(2023)Xue, Gao, Li, Qiao, Sun, Li, and
  Luo]{10.48550/arxiv.2211.08887}
Hongwei Xue, Peng Gao, Hongyang Li, Yu Qiao, Hao Sun, Houqiang Li, and Jiebo
  Luo.
\newblock Stare at what you see: {Masked} image modeling without
  reconstruction.
\newblock In \emph{Proceedings of the IEEE/CVF Conference on Computer Vision
  and Pattern Recognition}, pages 22732--22741, 2023.

\bibitem[Zhang et~al.(2022{\natexlab{a}})Zhang, Song, Oliveros, Fraczkiewicz,
  Legros, and Sandfeld]{Zhang_Song_Oliveros_Fraczkiewicz_Legros_Sandfeld_2022}
Chen Zhang, Hengxu Song, Daniela Oliveros, Anna Fraczkiewicz, Marc Legros, and
  Stefan Sandfeld.
\newblock Data-mining of in-situ {TEM} experiments: On the dynamics of
  dislocations in cocrfemnni alloys.
\newblock \emph{Acta Materialia}, 241:\penalty0 118394, 2022{\natexlab{a}}.

\bibitem[Zhang et~al.(2020)Zhang, Xie, Cai, Hu, Liu, Hong, and
  Tian]{zhang2020transfer}
Rui Zhang, Huimin Xie, Shuning Cai, Yong Hu, Guo-kun Liu, Wenjing Hong, and
  Zhong-qun Tian.
\newblock Transfer-learning-based raman spectra identification.
\newblock \emph{Journal of Raman Spectroscopy}, 51\penalty0 (1):\penalty0
  176--186, 2020.

\bibitem[Zhang et~al.(2022{\natexlab{b}})Zhang, Po, Xu, Liu, Ou, Zhao, and
  Yu]{10.1609/aaai.v36i3.20248}
Y. Zhang, L. Po, X. Xu, M. Liu, W. Ou, Y. Zhao, and W.~Y. Yu.
\newblock Contrastive spatio-temporal pretext learning for self-supervised
  video representation.
\newblock \emph{Proceedings of the AAAI Conference on Artificial Intelligence},
  36:\penalty0 3380--3389, 2022{\natexlab{b}}.

\bibitem[Zimmermann et~al.(2021)Zimmermann, Argus, and
  Brox]{zimmermann2021contrastive}
Christian Zimmermann, Max Argus, and Thomas Brox.
\newblock Contrastive representation learning for hand shape estimation.
\newblock In \emph{DAGM German Conference on Pattern Recognition}, pages
  250--264. Springer, 2021.

\end{thebibliography}
}

\clearpage
\onecolumn
\setcounter{page}{1}

\begin{Large}
\begin{center}
\textbf{Supplementary Material for Self-Supervised Learning with Generative Adversarial Networks for Electron Microscopy}

\author{Bashir Kazimi\textsuperscript{1}, Karina Ruzaeva\textsuperscript{1}, Stefan Sandfeld\textsuperscript{1,2}\\
{\textsuperscript{1}Forschungszentrum J\"ulich GmbH, J\"ulich, IAS-9, Germany}\\
{\textsuperscript{2}RWTH Aachen University, Aachen, Germany}\\
{\tt\small \{b.kazimi,k.ruzaeva,s.sandfeld\}@fz-juelich.de}
}
\end{center}
\end{Large}

This Supplementary Material includes example illustrations, tables and formulas that were either directly referred to in the main text or moved here due to page limits.
\subsection{Objective Function for cGANs}
The equation for the objective function in cGANs is as follows:

\begin{equation}
\begin{aligned}
    \mathcal{L}_\text{cGAN}(G, D) = & \mathbb{E}_{x,y} [\log D(x, y)] \\
    & + \mathbb{E}_{x,z} [\log(1 - D(x, G(x, z)))] \;,
\end{aligned}
\label{eq:ganloss}
\end{equation}

where the generator $G$ has the goal to minimize the loss and the discriminator $D$ attempts to maximize the loss. The generator $G$ is additionally trained to minimize the $L_1$ distance between its output and the original images
as given by the following equation:

\begin{equation}
    \mathcal{L}_L(G) = \mathbb{E}_{x,y,z} [\|y - G(x, z)\|_1] \;.
    \label{eq:l1}
\end{equation}

The overall objective function is
\begin{equation}
    G^* = \operatorname{arg} \min_G \max_D \mathcal{L}_\text{cGAN}(G, D) + \lambda \mathcal{L}_L(G)
    \label{eq:overallgan}
\end{equation}

where $\lambda$ controls how much weight should be given for the $L_1$ distance in the overall objective.
\subsection{Example Predictions for Semantic Segmentation}
\begin{figure}[!htb]
    \centering
    \includegraphics[width=.22\columnwidth]{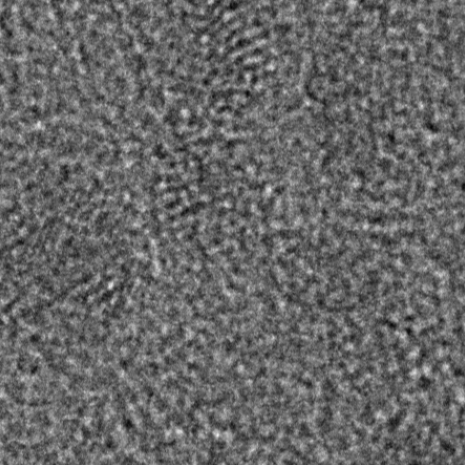}
    \includegraphics[width=.22\columnwidth]{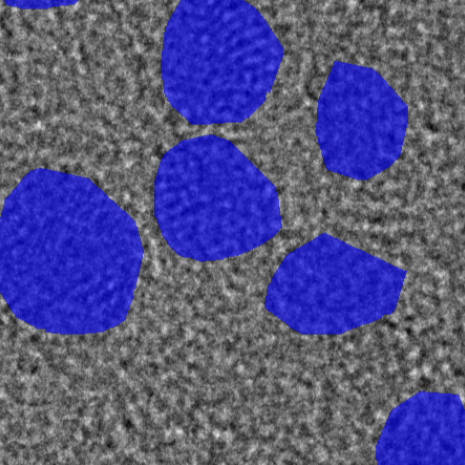}
    \includegraphics[width=.22\columnwidth]{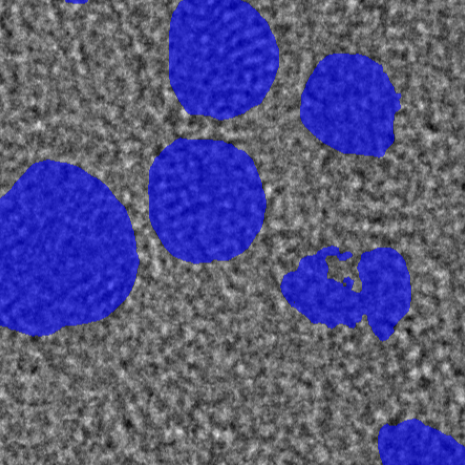}
    \includegraphics[width=.22\columnwidth]{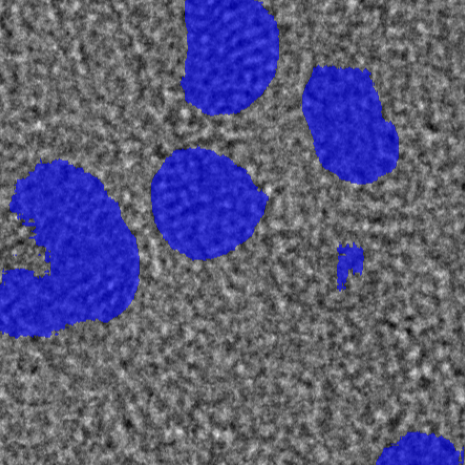}
    \caption{Left to write: original image, ground truth, prediction by HRNet P(100K) and U-Net\_4\_0 P(100K), respectively, for the Au2.2nm dataset.}
    \label{fig:au2p2segimages1}
\end{figure}

\begin{figure}[!htb]
    \centering
    \includegraphics[width=.22\columnwidth]{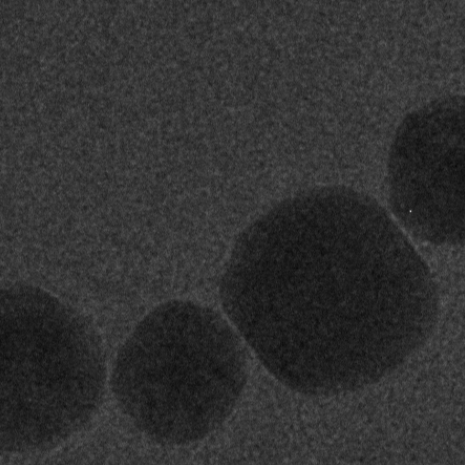}
    \includegraphics[width=.22\columnwidth]{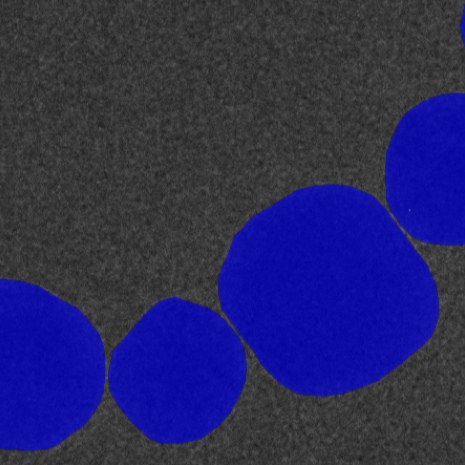}
    \includegraphics[width=.22\columnwidth]{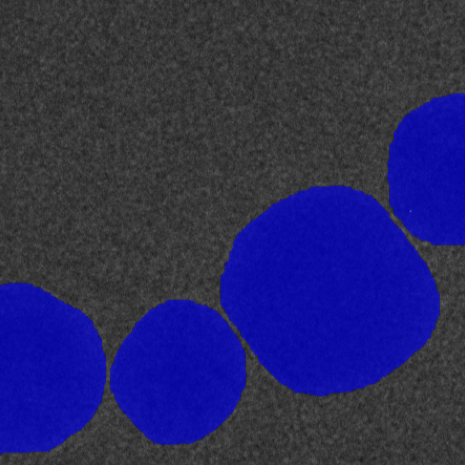}
    \includegraphics[width=.22\columnwidth]{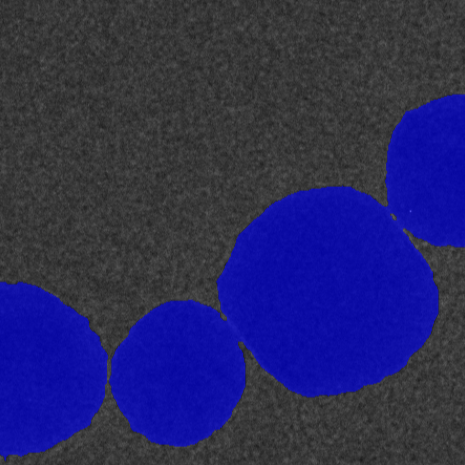}
    \caption{Left to write: original image, ground truth, prediction by HRNet P(100K) and U-Net\_4\_0 P(100K), respectively, for the Au20nm dataset.}
    \label{fig:au20titansegimages1}
\end{figure}

\begin{figure}[!htb]
    \centering
    \includegraphics[width=.22\columnwidth]{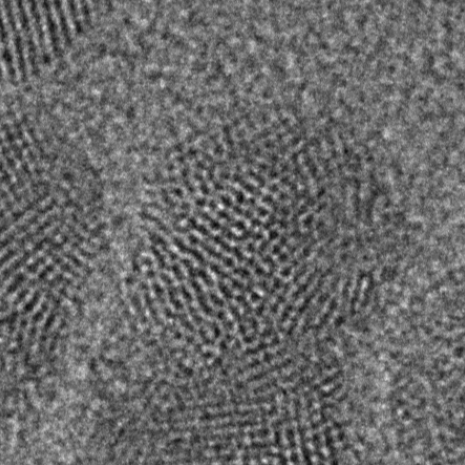}
    \includegraphics[width=.22\columnwidth]{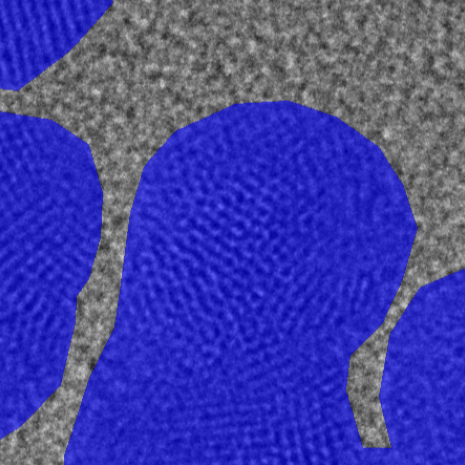}
    \includegraphics[width=.22\columnwidth]{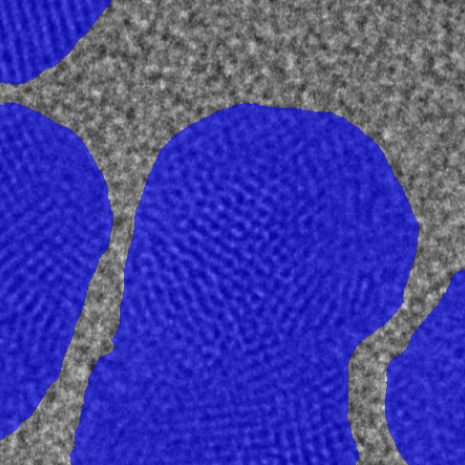}
    \includegraphics[width=.22\columnwidth]{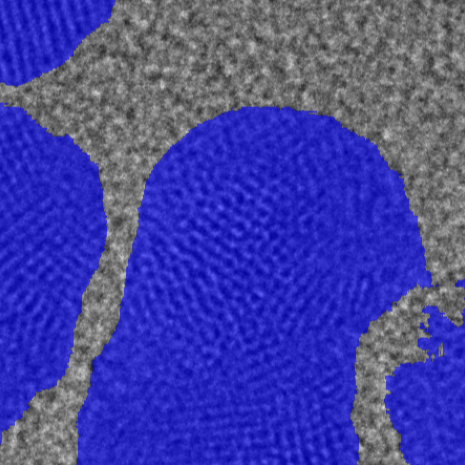}
    \caption{Left to write: original image, ground truth, prediction by HRNet P(\SI{100}{K}) and U-Net\_4\_0 P(\SI{100}{K}), respectively, for the Au5nm dataset.}
    \label{fig:au5titansegimages1}
\end{figure}

\begin{figure}[!htb]
    \centering
    \includegraphics[width=.23\columnwidth]{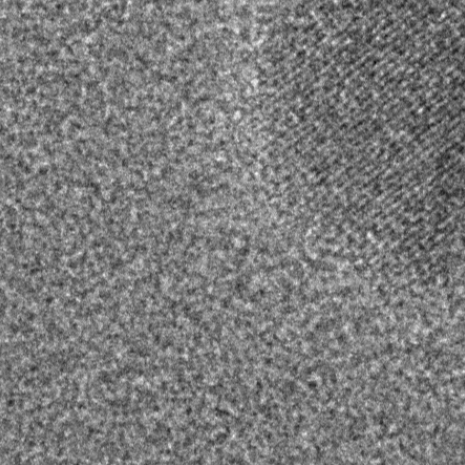}
    \includegraphics[width=.23\columnwidth]{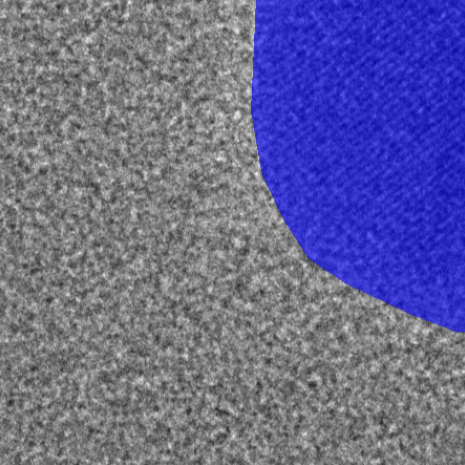}
    \includegraphics[width=.23\columnwidth]{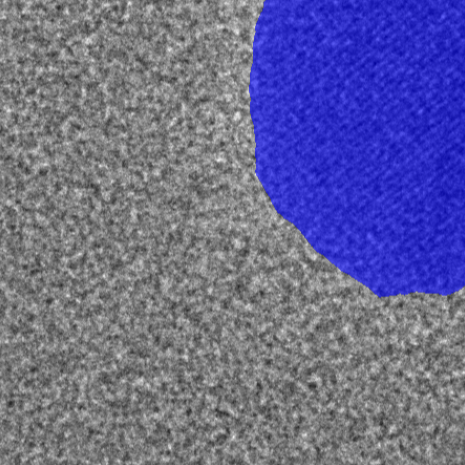}
    \includegraphics[width=.23\columnwidth]{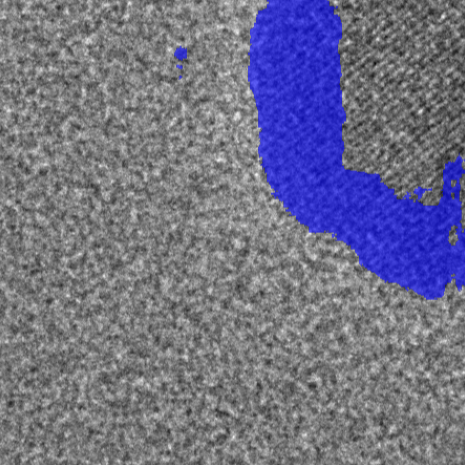}
    \caption{Left to write: original image, ground truth, prediction by HRNet P(\SI{100}{K}) and U-Net\_4\_0 P(\SI{100}{K}), respectively, for the Au10nm dataset.}
    \label{fig:au10titansegimages1}
\end{figure}

\newpage
\subsection{Validation Plots for the Pretraining}

\begin{figure*}[!ht]
    \centering



    \includegraphics[width=0.32\textwidth]{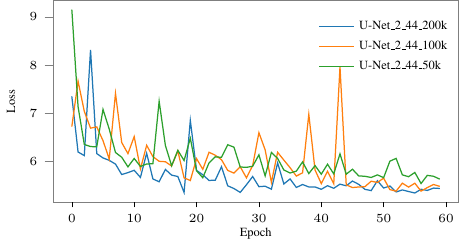}
    \includegraphics[width=0.32\textwidth]{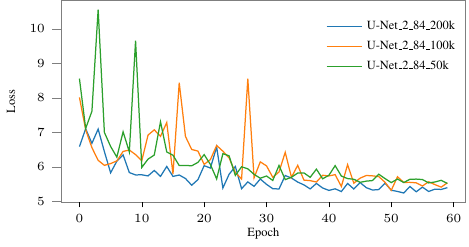}
    \includegraphics[width=0.32\textwidth]{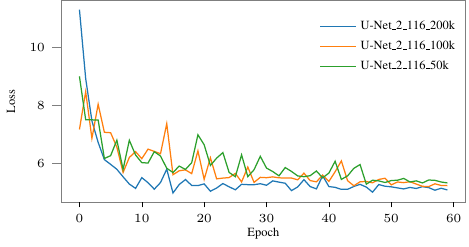}   

    \includegraphics[width=0.32\textwidth]{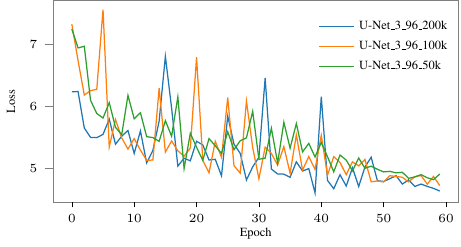}
    \includegraphics[width=0.32\textwidth]{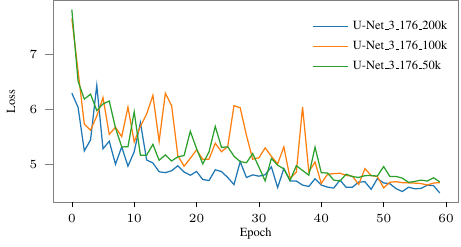}
    \includegraphics[width=0.32\textwidth]{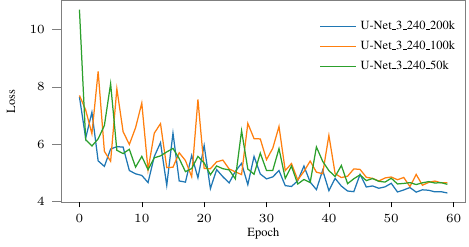}

    \includegraphics[width=0.32\textwidth]{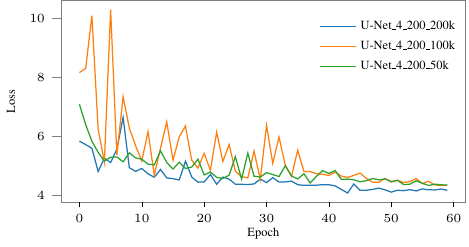}
    \includegraphics[width=0.32\textwidth]{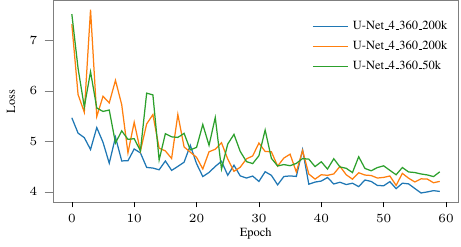}
    \caption{Validation $L_1$ loss for U-Nets (used as generator) of different sizes (with two, three, and four residual blocks) and receptive fields for three different dataset sizes.}
    \label{fig:denois_l1_Unets}
    
\end{figure*}
\newpage

\subsection{Quantitative Evaluations for Semantic Segmentation}
During training, the best model parameters are saved to disk after every 5 epochs. The following tables show model performance on the test datasets using the best models at the 5 epoch intervals. As observed in all tables, the higher performance is achieved in earlier epochs with fine-tuning while the random weight initialization leads to inferior performance or similar performance only towards the end of training.

\begin{table*}[!htb]
\caption{Comparison of segmentation Dice scores for different training methods (randomly initialized weights (R) and pretrained (P) with GANs on CEM500K using \SI{50}{K}, \SI{100}{K}, \SI{200}{K} images) on the Au5nmV1 test dataset. The experiments were performed with UNets of different sizes (with two, three, and four residual blocks) and receptive fields (three for each U-Net size) and HRNet.}
\label{au-table1}
\vskip 0.15in
\begin{center}
\begin{small}
\begin{sc}
\resizebox{\textwidth}{!}{
\begin{tabular}{lll|llllllllllll}
\multicolumn{3}{l|}{epochs} & \multicolumn{1}{c}{5} & \multicolumn{1}{c}{10} & \multicolumn{1}{c}{15} & \multicolumn{1}{c}{20} & \multicolumn{1}{c}{25} & \multicolumn{1}{c}{30} & \multicolumn{1}{c}{35} & \multicolumn{1}{c}{40} & \multicolumn{1}{c}{45} & \multicolumn{1}{c}{50} & \multicolumn{1}{c}{55} & \multicolumn{1}{c}{60} \\ \hline
\multicolumn{2}{l}{\multirow{4}{*}{\rotatebox[origin=c]{90}{HRNet}}} & R & 41.05 & 82.65 & 82.65 & 80.41 & 86.15 & 86.55 & 86.55 & 86.55 & 86.55 & 90.24 & 90.24 & 90.24 \\
\multicolumn{2}{l}{} & P(50k) & 86.37 & 91.03 & 91.41 & 91.41 & 91 & 91 & 92.2 & 91.97 & 91.97 & 91.97 & 91.97 & 91.97 \\
\multicolumn{2}{l}{} & P(100k) & 88.49 & 90.57 & 90.57 & 92.14 & 92.14 & 92.14 & 92.14 & 92.14 & 92.54 & 92.54 & 92.54 & 92.38 \\
\multicolumn{2}{l}{} & P(200k) & 83.86 & 89.71 & 89.71 & 91.41 & 91.41 & 91.41 & 91.41 & 91.41 & 92.3 & 92.3 & 92.3 & 92.07 \\ \hline
\multirow{12}{*}{\rotatebox[origin=c]{90}{U-Net 4 blocks}} & \multirow{4}{*}{\rotatebox[origin=c]{90}{424}} & R & 7.8 & 81.04 & 81.66 & 86.03 & 86.03 & 86.03 & 86.03 & 86.03 & 89.47 & 89.47 & 89.95 & 89.95 \\
 &  & P(50k) & 72.78 & 84.47 & 86.92 & 89.81 & 89.81 & 89.81 & 89.81 & 88.87 & 90.73 & 91.16 & 91.16 & 91.16 \\
 &  & P(100k) & 71.8 & 78.68 & 81.52 & 85.9 & 85.9 & 89 & 88.46 & 88.72 & 88.72 & 88.72 & 89.72 & 89.72 \\
 &  & P(200k) & 75.46 & 83.33 & 84.44 & 85.28 & 87.72 & 87.34 & 89.34 & 89.34 & 89.34 & 90.21 & 90.21 & 90.21 \\
 & \multirow{4}{*}{\rotatebox[origin=c]{90}{360}} & R & 0.01 & 84.65 & 84.65 & 84.65 & 84.65 & 84.65 & 84.65 & 84.65 & 84.65 & 91.07 & 91.07 & 91.07 \\
 &  & P(50k) & 68.33 & 78.12 & 83.2 & 84.48 & 85.83 & 85.83 & 85.83 & 85.83 & 85.83 & 89.09 & 90.08 & 90.08 \\
 &  & P(100k) & 71 & 78.74 & 81.66 & 86.07 & 86.07 & 86.07 & 88.11 & 88.11 & 89.21 & 90.41 & 90.41 & 90.41 \\
 &  & P(200k) & 70.97 & 81.86 & 78.92 & 87.15 & 88.79 & 89.45 & 89.45 & 89.45 & 89.45 & 89.45 & 89.45 & 90.86 \\
 & \multirow{4}{*}{\rotatebox[origin=c]{90}{200}} & R & 0.25 & 81.59 & 81.59 & 80.69 & 80.69 & 80.69 & 85.89 & 85.89 & 85.73 & 85.73 & 83.61 & 83.61 \\
 &  & P(50k) & 82.03 & 81.54 & 84.55 & 85.71 & 87.05 & 87.05 & 87.28 & 85.68 & 85.68 & 85.68 & 85.68 & 85.68 \\
 &  & P(100k) & 83 & 80.81 & 84.14 & 84.14 & 86.7 & 86.7 & 86.7 & 88.37 & 88.46 & 88.53 & 88.53 & 88.53 \\
 &  & P(200k) & 55.75 & 78.31 & 77.4 & 86.74 & 86.74 & 86.56 & 86.56 & 86.91 & 88.42 & 89.4 & 89 & 89 \\ \hline
\multirow{12}{*}{\rotatebox[origin=c]{90}{U-Net 3 blocks}} & \multirow{4}{*}{\rotatebox[origin=c]{90}{240}} & R & 0 & 72.59 & 69.97 & 69.97 & 82.7 & 84.41 & 84.41 & 84.41 & 84.41 & 86.82 & 86.76 & 89.82 \\
 &  & P(50k) & 57.2 & 81.81 & 85.23 & 87.38 & 87.38 & 87.38 & 87.9 & 87.9 & 87.9 & 87.94 & 87.94 & 87.94 \\
 &  & P(100k) & 50.9 & 77.5 & 77.5 & 85.12 & 85.12 & 85.96 & 85.96 & 84.85 & 86.79 & 86.79 & 86.79 & 86.79 \\
 &  & P(200k) & 49.13 & 75.12 & 84.69 & 84.69 & 86.08 & 88.4 & 89.08 & 89.08 & 89.08 & 89.13 & 89.13 & 89.73 \\
 & \multirow{4}{*}{\rotatebox[origin=c]{90}{176}} & R & 0 & 78.03 & 80.43 & 83.43 & 83.43 & 83.43 & 83.43 & 87.13 & 87.13 & 87.13 & 87.13 & 87.13 \\
 &  & P(50k) & 72.57 & 78.64 & 78.64 & 82.47 & 84.31 & 84.31 & 84.76 & 84.76 & 84.76 & 86.81 & 86.32 & 86.76 \\
 &  & P(100k) & 67.55 & 68.8 & 77.69 & 79.57 & 83.58 & 83.58 & 83.58 & 83.58 & 84.41 & 86.38 & 86.38 & 88.22 \\
 &  & P(200k) & 58.44 & 73.82 & 82.95 & 82.95 & 83.4 & 83.4 & 83.28 & 83.28 & 83.62 & 83.62 & 83.62 & 83.62 \\
 & \multirow{4}{*}{\rotatebox[origin=c]{90}{96}} & R & 0 & 67.22 & 75.79 & 80.41 & 80.41 & 82.8 & 82.8 & 82.8 & 84.41 & 84.41 & 84.41 & 84.41 \\
 &  & P(50k) & 65.65 & 76.38 & 81.28 & 81.3 & 82.5 & 82.86 & 84.69 & 84.69 & 84.69 & 83.96 & 83.96 & 83.96 \\
 &  & P(100k) & 74.22 & 76.38 & 80.19 & 82.1 & 82.73 & 83.31 & 83.85 & 83.85 & 83.85 & 83.85 & 83.85 & 84.37 \\
 &  & P(200k) & 71.55 & 81.88 & 82.04 & 82.04 & 81.5 & 81.62 & 83.65 & 83.65 & 83.65 & 84.59 & 84.59 & 84.27 \\ \hline
\multirow{12}{*}{\rotatebox[origin=c]{90}{U-Net 2 blocks}} & \multirow{4}{*}{\rotatebox[origin=c]{90}{116}} & R & 0 & 66.26 & 80.06 & 79.17 & 80.81 & 80.81 & 81.86 & 81.86 & 81.86 & 82.49 & 82.49 & 82.49 \\
 &  & P(50k) & 61.99 & 65.27 & 79.92 & 82.3 & 82.3 & 82.3 & 82.86 & 82.86 & 82.86 & 84.72 & 84.72 & 84.72 \\
 &  & P(100k) & 51.97 & 65.61 & 76.3 & 80.16 & 80.16 & 80.16 & 80.16 & 81.55 & 80.71 & 82.64 & 83.47 & 83.47 \\
 &  & P(200k) & 67.87 & 74.51 & 80.48 & 80.48 & 80.48 & 80.48 & 81.98 & 81.98 & 81.98 & 81.98 & 81.98 & 83.18 \\
 & \multirow{4}{*}{\rotatebox[origin=c]{90}{84}} & R & 0 & 73.4 & 75.97 & 75.97 & 78.77 & 80.77 & 80.77 & 82.82 & 82.41 & 82.41 & 82.41 & 81.76 \\
 &  & P(50k) & 66.69 & 75.04 & 77.83 & 82.53 & 83.05 & 83.75 & 83.75 & 83.75 & 83.72 & 83.26 & 83.18 & 83.99 \\
 &  & P(100k) & 70.28 & 76.14 & 78.62 & 81.33 & 81.44 & 81.44 & 81.44 & 81.44 & 81.44 & 82.52 & 82.52 & 82.63 \\
 &  & P(200k) & 63.88 & 77.09 & 79.08 & 82.87 & 82.87 & 82.73 & 82.73 & 83.6 & 83.6 & 83.6 & 83.6 & 83.6 \\
 & \multirow{4}{*}{\rotatebox[origin=c]{90}{44}} & R & 0 & 73.58 & 74.87 & 74.87 & 74.87 & 74.87 & 78 & 78 & 80.03 & 80.03 & 80.03 & 79.75 \\
 &  & P(50k) & 65.89 & 72.27 & 72.27 & 72.47 & 72.47 & 76.9 & 76.9 & 78.59 & 78.59 & 78.59 & 78.59 & 78.59 \\
 &  & P(100k) & 58.62 & 72.77 & 75.47 & 76.14 & 75.67 & 75.67 & 78.17 & 78.17 & 79.92 & 79.92 & 79.92 & 79.92 \\
 &  & P(200k) & 53.15 & 72.76 & 73.5 & 75.45 & 75.45 & 75.45 & 75.45 & 75.45 & 75.45 & 76.99 & 76.99 & 76.99
\end{tabular}%
}
\end{sc}
\end{small}
\end{center}
\vskip -0.1in
\end{table*}

\newpage
\begin{table*}[!htb]
\caption{Comparison of segmentation Dice scores for different training methods (randomly initialized weights (R) and pretrained (P) with GANs on CEM500K using \SI{50}{K}, \SI{100}{K}, \SI{200}{K} images) on the Au2.2nm test dataset. The experiments were performed with UNets of different sizes (with two, three, and four residual blocks) and receptive fields (three for each U-Net size) and HRNet.}
\label{au-table2}
\vskip 0.15in
\begin{center}
\begin{small}
\begin{sc}
\resizebox{\textwidth}{!}{
\begin{tabular}{lll|llllllllllll}
\multicolumn{3}{l|}{epochs} & \multicolumn{1}{c}{5} & \multicolumn{1}{c}{10} & \multicolumn{1}{c}{15} & \multicolumn{1}{c}{20} & \multicolumn{1}{c}{25} & \multicolumn{1}{c}{30} & \multicolumn{1}{c}{35} & \multicolumn{1}{c}{40} & \multicolumn{1}{c}{45} & \multicolumn{1}{c}{50} & \multicolumn{1}{c}{55} & \multicolumn{1}{c}{60} \\ \hline
\multicolumn{2}{l}{\multirow{4}{*}{\rotatebox[origin=c]{90}{HRNet}}} & R & 79.24 & 82.02 & 82.36 & 81.36 & 82.55 & 82.45 & 82.45 & 82.45 & 82.45 & 82.45 & 82.45 & 82.45 \\
\multicolumn{2}{l}{} & P(50k) & 81.57 & 82.16 & 83.1 & 83.1 & 83.94 & 83.94 & 83.94 & 83.94 & 83.94 & 83.94 & 83.94 & 83.94 \\
\multicolumn{2}{l}{} & P(100k) & 81.54 & 82.65 & 81.61 & 81.61 & 83.3 & 82.73 & 82.73 & 82.73 & 83.43 & 83.23 & 83.23 & 83.23 \\
\multicolumn{2}{l}{} & P(200k) & 81.14 & 82.11 & 82.11 & 82.11 & 82.11 & 82.11 & 82.11 & 82.11 & 82.11 & 82.11 & 82.11 & 82.11 \\ \hline
\multirow{12}{*}{\rotatebox[origin=c]{90}{U-Net 4 blocks}} & \multirow{4}{*}{\rotatebox[origin=c]{90}{424}} & R & 74.14 & 77.92 & 80.55 & 80.34 & 80.34 & 81.7 & 81.7 & 81.7 & 82.4 & 82.24 & 82.24 & 82.24 \\
 &  & P(50k) & 68.23 & 75.26 & 77.73 & 77.73 & 78.19 & 79.26 & 79.26 & 79.26 & 80.57 & 81.28 & 81.28 & 81.66 \\
 &  & P(100k) & 69.49 & 77.72 & 78.36 & 79.76 & 81.68 & 81.68 & 81.68 & 80.78 & 81.31 & 81.36 & 81.36 & 81.36 \\
 &  & P(200k) & 74.66 & 74.64 & 79.13 & 79.13 & 79.13 & 80.88 & 80.12 & 80.12 & 80.59 & 81.31 & 81.87 & 81.87 \\
 & \multirow{4}{*}{\rotatebox[origin=c]{90}{360}} & R & 0 & 77.17 & 80.42 & 80.42 & 80.42 & 80.73 & 81.41 & 81.41 & 81.36 & 81.94 & 81.94 & 82.81 \\
 &  & P(50k) & 73.96 & 77.54 & 77.54 & 79.01 & 81.16 & 81.16 & 81.65 & 81.65 & 81.36 & 80.94 & 81.89 & 81.89 \\
 &  & P(100k) & 73.47 & 74.21 & 77.43 & 81.3 & 79.15 & 82.05 & 82.05 & 82.05 & 82.05 & 81.55 & 81.55 & 81.55 \\
 &  & P(200k) & 67.11 & 79.06 & 79.28 & 79.28 & 80.1 & 80.1 & 81.18 & 79.69 & 79.69 & 79.69 & 81.39 & 80.98 \\
 & \multirow{4}{*}{\rotatebox[origin=c]{90}{200}} & R & 72.09 & 71.7 & 79.23 & 79.23 & 79.23 & 79.5 & 79.5 & 79.5 & 81.18 & 81.18 & 81.18 & 81.18 \\
 &  & P(50k) & 64.22 & 75.3 & 75.3 & 75.3 & 78.23 & 78.15 & 79.5 & 80.43 & 81.21 & 81.21 & 81.21 & 81.21 \\
 &  & P(100k) & 69.28 & 74.78 & 75.95 & 79.53 & 75.12 & 77.54 & 77.54 & 77.54 & 77.54 & 77.54 & 80.73 & 81.62 \\
 &  & P(200k) & 69.44 & 75.38 & 77.06 & 78.99 & 75.99 & 79.33 & 79.33 & 79.42 & 79.42 & 79.42 & 77.51 & 79.21 \\ \hline
\multirow{12}{*}{\rotatebox[origin=c]{90}{U-Net 3 blocks}} & \multirow{4}{*}{\rotatebox[origin=c]{90}{240}} & R & 0 & 70.35 & 77.22 & 79.21 & 79.63 & 79.63 & 79.63 & 81.18 & 81.08 & 82.07 & 82.07 & 82.07 \\
 &  & P(50k) & 73.25 & 71.77 & 75.24 & 78.41 & 78.41 & 78.41 & 78.87 & 81.36 & 79.65 & 79.65 & 79.65 & 79.65 \\
 &  & P(100k) & 71.05 & 69.95 & 73.86 & 77.06 & 79.4 & 79.4 & 79.4 & 80.82 & 80.9 & 80.9 & 80.9 & 81.81 \\
 &  & P(200k) & 66.61 & 70.65 & 73.29 & 74.62 & 75.53 & 76.75 & 78.01 & 78.83 & 78.83 & 78.83 & 78.83 & 78.16 \\
 & \multirow{4}{*}{\rotatebox[origin=c]{90}{176}} & R & 61.15 & 77.56 & 76.31 & 76.31 & 77.53 & 77.53 & 77.53 & 79.03 & 79.03 & 79.03 & 79.03 & 79.03 \\
 &  & P(50k) & 62.51 & 70.33 & 72.44 & 72.44 & 76.19 & 76.19 & 78.98 & 78.98 & 78.98 & 79.7 & 79.94 & 79.94 \\
 &  & P(100k) & 63.84 & 72.93 & 74.54 & 74.65 & 74.89 & 74.65 & 77.77 & 77.77 & 77.77 & 77.77 & 77.77 & 79.97 \\
 &  & P(200k) & 57.69 & 70.35 & 73.25 & 73.25 & 73.25 & 77.79 & 78.25 & 78.25 & 78.25 & 79.26 & 79.26 & 79.26 \\
 & \multirow{4}{*}{\rotatebox[origin=c]{90}{96}} & R & 0 & 66.01 & 68.19 & 68.19 & 67.04 & 67.97 & 67.97 & 67.97 & 69.94 & 69.94 & 70.2 & 70.2 \\
 &  & P(50k) & 57.88 & 65.46 & 67.07 & 67.07 & 69.48 & 69.48 & 69.48 & 67.67 & 67.67 & 67.67 & 71.09 & 71.09 \\
 &  & P(100k) & 58.8 & 58.8 & 64.66 & 64.66 & 64.66 & 64.66 & 64.66 & 69.28 & 69.28 & 69.28 & 70.25 & 70.25 \\
 &  & P(200k) & 54.24 & 62.23 & 62.23 & 67.87 & 67.87 & 64.66 & 64.66 & 66.3 & 68.44 & 68.26 & 68.26 & 67.95 \\ \hline
\multirow{12}{*}{\rotatebox[origin=c]{90}{U-Net 2 blocks}} & \multirow{4}{*}{\rotatebox[origin=c]{90}{116}} & R & 58.77 & 64.79 & 64.79 & 66.9 & 67.98 & 67.98 & 69.95 & 72.88 & 72.88 & 72.2 & 72.2 & 72.2 \\
 &  & P(50k) & 55.45 & 63.64 & 67.37 & 67.37 & 67.37 & 70.6 & 70.6 & 71.98 & 70.01 & 70.01 & 70.01 & 70.01 \\
 &  & P(100k) & 64.2 & 64.99 & 66.35 & 66.35 & 66.35 & 71.95 & 71.95 & 70.92 & 68.67 & 72.86 & 72.86 & 72.86 \\
 &  & P(200k) & 56.97 & 62.14 & 63.06 & 66.4 & 67.59 & 66.49 & 66.49 & 66.49 & 67.76 & 67.76 & 68.35 & 68.35 \\
 & \multirow{4}{*}{\rotatebox[origin=c]{90}{84}} & R & 0 & 0 & 57.5 & 64.67 & 64.67 & 64.67 & 68.2 & 68.2 & 68.29 & 68.29 & 68.29 & 67.8 \\
 &  & P(50k) & 58.08 & 65.29 & 63.2 & 65.06 & 64.9 & 65.23 & 66.48 & 66.48 & 66.48 & 66.48 & 66.96 & 66.96 \\
 &  & P(100k) & 52.37 & 61.42 & 62.52 & 62.52 & 62.52 & 66.07 & 66.07 & 66.07 & 66.07 & 66.07 & 66.07 & 66.07 \\
 &  & P(200k) & 53.46 & 62.37 & 62.37 & 62.37 & 62.37 & 65.81 & 64.07 & 66.83 & 64.97 & 64.97 & 64.97 & 64.97 \\
 & \multirow{4}{*}{\rotatebox[origin=c]{90}{84}} & R & 0 & 0 & 0 & 51.89 & 59.29 & 60.1 & 59.4 & 59.4 & 59.4 & 59.4 & 61.18 & 61.18 \\
 &  & P(50k) & 46.56 & 46.56 & 48.67 & 56.55 & 53.71 & 53.71 & 53.71 & 53.71 & 53.71 & 57.93 & 59.3 & 59.3 \\
 &  & P(100k) & 47.67 & 55.09 & 55.86 & 55.86 & 57.37 & 57.37 & 57.37 & 57.37 & 57.37 & 59.96 & 59.96 & 59.96 \\
 &  & P(200k) & 41.06 & 47.32 & 50.83 & 50.92 & 52.07 & 52.38 & 56.32 & 56.63 & 58.46 & 58.46 & 58.46 & 58.46
\end{tabular}%
}
\end{sc}
\end{small}
\end{center}
\vskip -0.1in
\end{table*}
\newpage

\begin{table*}[!htb]
\caption{Comparison of segmentation Dice scores (in \%) for different training methods (randomly initialized weights (R) and pretrained (P) with GANs on CEM500K using \SI{50}{K}, \SI{100}{K}, \SI{200}{K} images) on the Au5nm test datasets. The experiments were performed with UNets of different sizes (with two, three, and four residual blocks) and receptive fields (three for each U-Net size) and HRNet.}
\label{au-table3}
\vskip 0.15in
\begin{center}
\begin{small}
\begin{sc}
\resizebox{\textwidth}{!}{
\begin{tabular}{lll|llllllllllll}
\multicolumn{3}{l|}{epochs} & \multicolumn{1}{c}{5} & \multicolumn{1}{c}{10} & \multicolumn{1}{c}{15} & \multicolumn{1}{c}{20} & \multicolumn{1}{c}{25} & \multicolumn{1}{c}{30} & \multicolumn{1}{c}{35} & \multicolumn{1}{c}{40} & \multicolumn{1}{c}{45} & \multicolumn{1}{c}{50} & \multicolumn{1}{c}{55} & \multicolumn{1}{c}{60} \\ \hline
\multicolumn{2}{l}{\multirow{4}{*}{\rotatebox[origin=c]{90}{HRNet}}} & R & 86.29 & 91.12 & 92.61 & 93.29 & 93.72 & 94.43 & 94.12 & 94.12 & 94.44 & 94.68 & 94.72 & 94.76 \\
\multicolumn{2}{l}{} & P(50k) & 92.51 & 94.46 & 94.35 & 94.56 & 94.56 & 94.54 & 94.6 & 94.6 & 94.62 & 94.62 & 94.56 & 94.56 \\
\multicolumn{2}{l}{} & P(100k) & 93.74 & 94.24 & 94.24 & 94.24 & 94.25 & 94.4 & 94.4 & 94.4 & 94.53 & 94.53 & 94.79 & 94.6 \\
\multicolumn{2}{l}{} & P(200k) & 93.27 & 93.9 & 94.05 & 94.5 & 94.63 & 94.63 & 94.63 & 94.63 & 94.57 & 94.57 & 94.57 & 94.57 \\ \hline
\multirow{12}{*}{\rotatebox[origin=c]{90}{U-Net 4 blocks}} & \multirow{4}{*}{\rotatebox[origin=c]{90}{424}} & R & 88.69 & 91.98 & 92.55 & 93.67 & 93.11 & 93.81 & 94.16 & 94.16 & 94.21 & 94.21 & 94.21 & 94.36 \\
 &  & P(50k) & 88.29 & 90.08 & 92.16 & 92.13 & 93.34 & 93.34 & 93.78 & 93.78 & 93.78 & 93.74 & 93.81 & 93.81 \\
 &  & P(100k) & 83.95 & 90.59 & 91.31 & 87.77 & 92.8 & 93.7 & 93.7 & 93.72 & 93.67 & 93.67 & 94.18 & 94.18 \\
 &  & P(200k) & 79.55 & 89.32 & 91.88 & 92.26 & 90.47 & 92.94 & 93.26 & 93.35 & 92.15 & 92.15 & 93.68 & 93.72 \\
 & \multirow{4}{*}{\rotatebox[origin=c]{90}{360}} & R & 86.82 & 91.67 & 92.99 & 93.63 & 93.83 & 94.16 & 94.29 & 93.87 & 93.87 & 94.19 & 94.19 & 94.19 \\
 &  & P(50k) & 83.64 & 89.7 & 89.7 & 92.39 & 92.67 & 92.67 & 92.82 & 93.9 & 93.9 & 93.9 & 93.79 & 94.28 \\
 &  & P(100k) & 87.61 & 90.61 & 90.61 & 92.22 & 92.22 & 92.22 & 92.22 & 93.56 & 93.56 & 93.56 & 94.08 & 94.08 \\
 &  & P(200k) & 88.58 & 90.44 & 91.02 & 92.62 & 92.62 & 93.47 & 93.47 & 94.07 & 94.07 & 94.07 & 94.07 & 94.07 \\
 & \multirow{4}{*}{\rotatebox[origin=c]{90}{200}} & R & 80.25 & 78.81 & 82.84 & 82.84 & 87.28 & 87.04 & 83.93 & 84.18 & 84.18 & 84.18 & 84.18 & 90.53 \\
 &  & P(50k) & 78.6 & 86.96 & 89.26 & 91.27 & 91.27 & 92.18 & 92.18 & 91.45 & 92.85 & 92.2 & 92.2 & 93.16 \\
 &  & P(100k) & 80.63 & 86.14 & 73.63 & 75.68 & 75.68 & 91.96 & 91.96 & 91.96 & 91.32 & 91.82 & 91.82 & 91.82 \\
 &  & P(200k) & 81.92 & 85.52 & 88.2 & 91.22 & 91.34 & 92.19 & 92.19 & 93.16 & 93.16 & 93.16 & 93.34 & 93.34 \\ \hline
\multirow{12}{*}{\rotatebox[origin=c]{90}{U-Net 3 blocks}} & \multirow{4}{*}{\rotatebox[origin=c]{90}{240}} & R & 0 & 87.23 & 84.49 & 90.06 & 89.14 & 90.35 & 89.05 & 88.55 & 88.55 & 87.3 & 87.3 & 87.3 \\
 &  & P(50k) & 77.28 & 83.27 & 86.52 & 90.3 & 90.9 & 90.9 & 90.9 & 90.9 & 92.79 & 92.79 & 92.1 & 92.1 \\
 &  & P(100k) & 77.26 & 81.34 & 85.51 & 85.51 & 85.53 & 85.53 & 85.53 & 85.53 & 85.53 & 92.41 & 92.63 & 89.37 \\
 &  & P(200k) & 79.95 & 81.54 & 81.54 & 81.54 & 89.23 & 89.23 & 89.92 & 89.92 & 91.87 & 91.87 & 91.87 & 92.64 \\
 & \multirow{4}{*}{\rotatebox[origin=c]{90}{176}} & R & 71.12 & 62.29 & 82.12 & 82.12 & 77.2 & 72.48 & 72.48 & 72.48 & 72.48 & 72.48 & 72.48 & 74.18 \\
 &  & P(50k) & 77.35 & 77.35 & 81.69 & 84.35 & 84.35 & 88.77 & 88.77 & 81.7 & 81.7 & 81.7 & 83.43 & 83.43 \\
 &  & P(100k) & 74.26 & 84.6 & 84.6 & 88.68 & 89.31 & 87.76 & 88.43 & 88.43 & 88.43 & 88.43 & 88.43 & 91.3 \\
 &  & P(200k) & 77.37 & 83.23 & 85.75 & 87.47 & 89.38 & 89.37 & 89.37 & 89.37 & 88.33 & 91.4 & 91.4 & 91.4 \\
 & \multirow{4}{*}{\rotatebox[origin=c]{90}{96}} & R & 75.95 & 82.46 & 79.72 & 79.72 & 78.78 & 78.78 & 78.78 & 78.78 & 85.45 & 87.21 & 87.21 & 82.6 \\
 &  & P(50k) & 80.55 & 82.35 & 82.76 & 82.76 & 82.76 & 83.91 & 83.13 & 83.13 & 83.13 & 83.13 & 85.67 & 84.31 \\
 &  & P(100k) & 73.9 & 79.13 & 79.13 & 78.79 & 78.79 & 78.79 & 78.79 & 78.79 & 80.35 & 80.35 & 80.35 & 80.35 \\
 &  & P(200k) & 78.6 & 75.64 & 76.25 & 76.25 & 76.25 & 76.25 & 76.25 & 84.86 & 84.13 & 84.13 & 84.13 & 84.13 \\ \hline
\multirow{12}{*}{\rotatebox[origin=c]{90}{U-Net 2 blocks}} & \multirow{4}{*}{\rotatebox[origin=c]{90}{116}} & R & 80.66 & 74.92 & 74.49 & 74.49 & 74.49 & 74.49 & 82.26 & 82.26 & 76.51 & 76.51 & 76.51 & 76.51 \\
 &  & P(50k) & 79.88 & 79.88 & 78.52 & 78.52 & 78.52 & 78.52 & 77.69 & 77.69 & 77.69 & 77.69 & 77.69 & 77.69 \\
 &  & P(100k) & 73.76 & 78.17 & 78.17 & 82.27 & 81.74 & 81.74 & 82.46 & 71.87 & 71.87 & 71.87 & 71.87 & 71.87 \\
 &  & P(200k) & 76.25 & 79.29 & 82.11 & 81.23 & 81.23 & 81.23 & 81.23 & 81.23 & 81.23 & 81.23 & 84.09 & 84.09 \\
 & \multirow{4}{*}{\rotatebox[origin=c]{90}{84}} & R & 67.13 & 80.18 & 78.57 & 78.57 & 78.57 & 78.57 & 76.39 & 76.39 & 76.39 & 76.39 & 78.6 & 78.6 \\
 &  & P(50k) & 80.19 & 80.19 & 80.19 & 80.19 & 75.73 & 78.29 & 78.29 & 76.89 & 76.89 & 76.89 & 76.89 & 76.89 \\
 &  & P(100k) & 77.4 & 80.85 & 80.85 & 81.67 & 81.67 & 82.81 & 82.81 & 82.81 & 83.03 & 83.03 & 83.03 & 81.76 \\
 &  & P(200k) & 64.98 & 79.31 & 79.31 & 79.31 & 79.31 & 81.26 & 81.26 & 81.26 & 82.14 & 82.14 & 82.14 & 82.45 \\
 & \multirow{4}{*}{\rotatebox[origin=c]{90}{84}} & R & 70.5 & 75.06 & 75.06 & 76.63 & 76.1 & 76.1 & 76.1 & 76.1 & 76.1 & 76.1 & 77.83 & 77.83 \\
 &  & P(50k) & 65.87 & 72.88 & 72.88 & 70.8 & 70.8 & 70.8 & 70.8 & 70.8 & 70.8 & 70.8 & 74.39 & 74.39 \\
 &  & P(100k) & 68.77 & 70.86 & 73 & 75.13 & 76.49 & 76.47 & 76.47 & 76.47 & 76.13 & 77.11 & 77.11 & 77.11 \\
 &  & P(200k) & 60.94 & 63.7 & 69.83 & 73.56 & 73.56 & 73.66 & 73.66 & 73.66 & 73.66 & 73.66 & 75.25 & 76.72
\end{tabular}%
}
\end{sc}
\end{small}
\end{center}
\vskip -0.1in
\end{table*}
\newpage

\begin{table*}[!htb]
\caption{Comparison of segmentation Dice scores (in \%) for different training methods (randomly initialized weights (R) and pretrained (P) with GANs on CEM500K using \SI{50}{K}, \SI{100}{K}, \SI{200}{K} images) on the Au10nm test dataset. The experiments were performed with UNets of different sizes (with two, three, and four residual blocks) and receptive fields (three for each U-Net size) and HRNet.}
\label{au-table4}
\vskip 0.15in
\begin{center}
\begin{small}
\begin{sc}
\resizebox{\textwidth}{!}{
\begin{tabular}{lll|llllllllllll}
\multicolumn{3}{l|}{epochs} & \multicolumn{1}{c}{5} & \multicolumn{1}{c}{10} & \multicolumn{1}{c}{15} & \multicolumn{1}{c}{20} & \multicolumn{1}{c}{25} & \multicolumn{1}{c}{30} & \multicolumn{1}{c}{35} & \multicolumn{1}{c}{40} & \multicolumn{1}{c}{45} & \multicolumn{1}{c}{50} & \multicolumn{1}{c}{55} & \multicolumn{1}{c}{60} \\ \hline
\multicolumn{2}{l}{\multirow{4}{*}{\rotatebox[origin=c]{90}{HRNet}}} & R & 82.66 & 92.47 & 92.36 & 95.13 & 95.13 & 95.29 & 95.7 & 95.7 & 95.7 & 95.7 & 95.7 & 96.38 \\
\multicolumn{2}{l}{} & P(50k) & 94.66 & 95.17 & 96.53 & 96.39 & 96.79 & 96.96 & 96.91 & 97.19 & 97.19 & 97.28 & 97.34 & 97.25 \\
\multicolumn{2}{l}{} & P(100k) & 94.82 & 96.74 & 96.74 & 97.08 & 97.08 & 97.08 & 97.08 & 97.08 & 97.08 & 97.08 & 97.08 & 97.08 \\
\multicolumn{2}{l}{} & P(200k) & 94.33 & 96.07 & 96.07 & 96.56 & 96.56 & 96.56 & 96.99 & 97.43 & 97.43 & 97.32 & 97.36 & 97.36 \\ \hline
\multirow{12}{*}{\rotatebox[origin=c]{90}{U-Net 4 blocks}} & \multirow{4}{*}{\rotatebox[origin=c]{90}{424}} & R & 86.01 & 90.11 & 92.37 & 94.45 & 94.61 & 95.93 & 95.93 & 95.82 & 95.82 & 96.5 & 96.5 & 96.6 \\
 &  & P(50k) & 89.89 & 92 & 92.52 & 94.05 & 94.41 & 95.07 & 95.07 & 95.25 & 95.88 & 96.09 & 96.09 & 96.27 \\
 &  & P(100k) & 86.18 & 90.24 & 92.06 & 92.06 & 93.68 & 93.2 & 93.83 & 94.62 & 94.74 & 95.15 & 95.36 & 95.36 \\
 &  & P(200k) & 85.56 & 90.86 & 91.47 & 92.45 & 93.78 & 93.78 & 93.78 & 94.12 & 94.12 & 95.31 & 95.06 & 95.06 \\
 & \multirow{4}{*}{\rotatebox[origin=c]{90}{360}} & R & 83.15 & 90.79 & 90.79 & 93.14 & 92.67 & 93.53 & 94.92 & 94.92 & 94.92 & 95.37 & 95.37 & 96.02 \\
 &  & P(50k) & 83.66 & 90.32 & 85.78 & 93.26 & 94.12 & 94.12 & 94.17 & 95.14 & 95.96 & 95.74 & 95.74 & 95.74 \\
 &  & P(100k) & 86.72 & 91.33 & 91.89 & 93.7 & 94.29 & 94.29 & 94.29 & 94.29 & 94.62 & 95.6 & 95.6 & 95.6 \\
 &  & P(200k) & 85.99 & 90.75 & 92.05 & 92.52 & 93.68 & 93.68 & 94.15 & 94.31 & 94.31 & 94.31 & 94.9 & 94.62 \\
 & \multirow{4}{*}{\rotatebox[origin=c]{90}{200}} & R & 78.11 & 84.92 & 91.39 & 94.36 & 94.16 & 94.32 & 94.32 & 95.83 & 95.83 & 95.83 & 95.26 & 96.04 \\
 &  & P(50k) & 82.69 & 89.68 & 91.5 & 92.08 & 93.42 & 94.61 & 94.86 & 95 & 95 & 95 & 95 & 95 \\
 &  & P(100k) & 90.1 & 90.64 & 92.12 & 92.12 & 94.07 & 94.07 & 93.79 & 93.79 & 94.93 & 94.93 & 95.78 & 94.11 \\
 &  & P(200k) & 89.46 & 90.59 & 90.59 & 92.56 & 92.56 & 93.06 & 93.06 & 93.82 & 93.82 & 93.82 & 93.82 & 94.74 \\ \hline
\multirow{12}{*}{\rotatebox[origin=c]{90}{U-Net 3 blocks}} & \multirow{4}{*}{\rotatebox[origin=c]{90}{240}} & R & 78.95 & 84.47 & 89.45 & 88.39 & 91.73 & 91.73 & 93.79 & 94.16 & 94.16 & 94.16 & 94.16 & 94.16 \\
 &  & P(50k) & 89.67 & 87.54 & 89.51 & 89.51 & 91.54 & 92.5 & 93.31 & 93.31 & 93.83 & 93.83 & 93.83 & 94.8 \\
 &  & P(100k) & 77.97 & 88.65 & 88.65 & 88.42 & 93.28 & 93.28 & 92.82 & 92.82 & 92.82 & 93.91 & 93.91 & 93.91 \\
 &  & P(200k) & 84.82 & 88.71 & 89.47 & 91.25 & 91.25 & 91.25 & 92.44 & 93.23 & 93.23 & 93.23 & 93.23 & 93.23 \\
 & \multirow{4}{*}{\rotatebox[origin=c]{90}{176}} & R & 80.21 & 88.64 & 90.48 & 92.11 & 92.11 & 94.28 & 93.97 & 93.97 & 93.97 & 94.95 & 94.95 & 95.44 \\
 &  & P(50k) & 80.09 & 87.77 & 87.77 & 91.06 & 91.06 & 91.7 & 93.42 & 92.95 & 92.43 & 92.43 & 92.43 & 92.43 \\
 &  & P(100k) & 80.63 & 85.92 & 89.67 & 90.96 & 91.17 & 92.16 & 92.16 & 92.16 & 92.16 & 93.24 & 93 & 93.98 \\
 &  & P(200k) & 83.75 & 86.88 & 91.33 & 91.92 & 91.9 & 91.9 & 92.86 & 92.86 & 92.86 & 93.91 & 93.91 & 94.58 \\
 & \multirow{4}{*}{\rotatebox[origin=c]{90}{96}} & R & 74.58 & 87.31 & 90.56 & 90.56 & 91.12 & 92.12 & 92.12 & 92.12 & 92.12 & 92.7 & 92.99 & 92.99 \\
 &  & P(50k) & 86.37 & 86.37 & 86.37 & 86.37 & 89.31 & 90.6 & 90.6 & 90.6 & 91.01 & 90.21 & 90.21 & 91.35 \\
 &  & P(100k) & 82.64 & 82.95 & 86.43 & 88.35 & 88.45 & 88.45 & 88.45 & 89.22 & 90.65 & 90.65 & 90.65 & 90.77 \\
 &  & P(200k) & 80.26 & 84.84 & 84.84 & 88.34 & 88.34 & 88.34 & 89.89 & 89.89 & 89.49 & 89.6 & 91.31 & 91.6 \\ \hline
\multirow{12}{*}{\rotatebox[origin=c]{90}{U-Net 2 blocks}} & \multirow{4}{*}{\rotatebox[origin=c]{90}{116}} & R & 77.99 & 88.37 & 88.87 & 88.87 & 88.87 & 90.74 & 90.74 & 90.74 & 90.74 & 90.74 & 92.62 & 91.97 \\
 &  & P(50k) & 84.63 & 88.12 & 89.26 & 89.26 & 89.26 & 89.26 & 91.23 & 91.23 & 91.23 & 91.23 & 91.23 & 91.23 \\
 &  & P(100k) & 81.11 & 89.75 & 89.75 & 89.75 & 89.75 & 89.75 & 89.75 & 89.75 & 91.26 & 91.26 & 90.52 & 90.52 \\
 &  & P(200k) & 74.59 & 83.77 & 85.98 & 87.29 & 87.29 & 89.49 & 89.49 & 90.05 & 90.05 & 90.05 & 90.54 & 91.53 \\
 & \multirow{4}{*}{\rotatebox[origin=c]{90}{84}} & R & 67.42 & 79.87 & 83.86 & 89.3 & 89.03 & 89.97 & 89.97 & 91.03 & 91.12 & 91.12 & 91.12 & 90.93 \\
 &  & P(50k) & 84.9 & 84.41 & 84.41 & 84.41 & 86.84 & 86.84 & 86.84 & 86.84 & 86.84 & 89.53 & 89.53 & 89.53 \\
 &  & P(100k) & 83.1 & 83.1 & 83.1 & 88.26 & 88.26 & 87.07 & 87.07 & 87.07 & 89.7 & 89.7 & 89.7 & 91.31 \\
 &  & P(200k) & 77.11 & 83.31 & 83.31 & 83.31 & 83.31 & 83.31 & 88.58 & 89.34 & 89.34 & 89.34 & 89.34 & 89.34 \\
 & \multirow{4}{*}{\rotatebox[origin=c]{90}{84}} & R & 68.75 & 71.89 & 77.94 & 80.52 & 84.14 & 84.14 & 84.14 & 85.66 & 85.66 & 86.39 & 86.39 & 86.68 \\
 &  & P(50k) & 75.08 & 75.08 & 75.08 & 77.74 & 77.74 & 77.74 & 77.74 & 77.74 & 83.8 & 83.8 & 83.8 & 83.8 \\
 &  & P(100k) & 79.16 & 79.16 & 79.16 & 79.16 & 79.16 & 83.66 & 83.66 & 83.66 & 83.66 & 83.66 & 83.66 & 83.66 \\
 &  & P(200k) & 67.11 & 71.81 & 71.81 & 74.74 & 75.45 & 75.45 & 75.45 & 78.73 & 78.73 & 78.36 & 78.36 & 78.36
\end{tabular}%
}
\end{sc}
\end{small}
\end{center}
\vskip -0.1in
\end{table*}
\newpage

\begin{table*}[!htb]
\caption{Comparison of segmentation Dice scores (in \%) for different training methods (randomly initialized weights (R) and pretrained (P) with GANs on CEM500K using \SI{50}{K}, \SI{100}{K}, \SI{200}{K} images) on the Au20nm test dataset. The experiments were performed with UNets of different sizes (with two, three, and four residual blocks) and receptive fields (three for each U-Net size) and HRNet.}
\label{au-table5}
\vskip 0.15in
\begin{center}
\begin{small}
\begin{sc}
\resizebox{\textwidth}{!}{
\begin{tabular}{lll|llllllllllll}
\multicolumn{3}{l|}{epochs} & \multicolumn{1}{c}{5} & \multicolumn{1}{c}{10} & \multicolumn{1}{c}{15} & \multicolumn{1}{c}{20} & \multicolumn{1}{c}{25} & \multicolumn{1}{c}{30} & \multicolumn{1}{c}{35} & \multicolumn{1}{c}{40} & \multicolumn{1}{c}{45} & \multicolumn{1}{c}{50} & \multicolumn{1}{c}{55} & \multicolumn{1}{c}{60} \\ \hline
\multicolumn{2}{l}{\multirow{4}{*}{\rotatebox[origin=c]{90}{HRNet}}} & R & 89.27 & 92.19 & 95.61 & 96.65 & 97.37 & 97.69 & 97.69 & 97.69 & 97.87 & 98.11 & 98.3 & 98.37 \\
\multicolumn{2}{l}{} & P(50k) & 97.8 & 98.17 & 98.45 & 98.48 & 98.56 & 98.5 & 98.61 & 98.63 & 98.65 & 98.67 & 98.65 & 98.67 \\
\multicolumn{2}{l}{} & P(100k) & 97.88 & 98.24 & 98.35 & 98.48 & 98.52 & 98.56 & 98.59 & 98.64 & 98.67 & 98.7 & 98.67 & 98.71 \\
\multicolumn{2}{l}{} & P(200k) & 98.08 & 98.39 & 98.53 & 98.53 & 98.53 & 98.53 & 98.6 & 98.6 & 98.62 & 98.62 & 98.67 & 98.69 \\ \hline
\multirow{12}{*}{\rotatebox[origin=c]{90}{U-Net 4 blocks}} & \multirow{4}{*}{\rotatebox[origin=c]{90}{424}} & R & 92.18 & 92.18 & 96.12 & 96.99 & 97.34 & 97.58 & 97.64 & 97.64 & 97.87 & 97.87 & 97.87 & 97.97 \\
 &  & P(50k) & 95.19 & 96.79 & 97.44 & 97.7 & 97.89 & 98.02 & 98.12 & 98.12 & 98.06 & 98.27 & 98.27 & 98.27 \\
 &  & P(100k) & 96.06 & 97.05 & 97.51 & 97.63 & 97.73 & 97.87 & 98.01 & 98.06 & 98.17 & 98.22 & 98.26 & 98.26 \\
 &  & P(200k) & 94.14 & 96.22 & 97.4 & 97.84 & 97.92 & 98.01 & 98 & 98.13 & 98.1 & 98.23 & 98.27 & 98.27 \\
 & \multirow{4}{*}{\rotatebox[origin=c]{90}{360}} & R & 76.57 & 91.93 & 94.55 & 95.1 & 96.7 & 97.26 & 97.26 & 97.26 & 97.26 & 97.45 & 97.69 & 97.7 \\
 &  & P(50k) & 95.43 & 96.78 & 97.4 & 97.76 & 97.88 & 98.01 & 98.01 & 98.13 & 98.13 & 98.13 & 98.13 & 98.11 \\
 &  & P(100k) & 94.58 & 96.78 & 97.2 & 97.72 & 97.87 & 97.91 & 98.02 & 98.11 & 98.14 & 98.17 & 98.18 & 98.26 \\
 &  & P(200k) & 95.24 & 96.92 & 97.56 & 97.78 & 97.94 & 97.94 & 97.99 & 98.14 & 98.14 & 98.25 & 98.25 & 98.25 \\
 & \multirow{4}{*}{\rotatebox[origin=c]{90}{200}} & R & 0 & 0 & 0 & 96.58 & 97.56 & 97.56 & 97.56 & 97.72 & 97.95 & 98.06 & 98.1 & 98.13 \\
 &  & P(50k) & 95.23 & 95.98 & 97.55 & 97.35 & 97.61 & 97.61 & 98.03 & 98.14 & 98.23 & 98.27 & 98.27 & 98.27 \\
 &  & P(100k) & 93.89 & 96.21 & 97.66 & 97.71 & 97.89 & 98.09 & 98.09 & 98.13 & 98.18 & 98.18 & 98.23 & 98.3 \\
 &  & P(200k) & 94.6 & 96.53 & 97.21 & 97.63 & 97.84 & 97.84 & 98.08 & 98.14 & 98.14 & 98.14 & 98.25 & 98.31 \\ \hline
\multirow{12}{*}{\rotatebox[origin=c]{90}{U-Net 3 blocks}} & \multirow{4}{*}{\rotatebox[origin=c]{90}{240}} & R & 93.82 & 95.65 & 96.49 & 97.25 & 97.57 & 97.76 & 97.91 & 97.93 & 98.14 & 98.23 & 98.23 & 98.32 \\
 &  & P(50k) & 95.82 & 96.38 & 96.95 & 97.37 & 97.71 & 97.68 & 97.88 & 97.88 & 97.88 & 98.04 & 98.04 & 98.14 \\
 &  & P(100k) & 93.26 & 95.84 & 97.13 & 97.39 & 97.41 & 97.63 & 97.72 & 97.82 & 97.82 & 97.88 & 97.88 & 98.01 \\
 &  & P(200k) & 91.91 & 95.42 & 96.97 & 96.97 & 97.12 & 97.7 & 97.73 & 97.78 & 97.94 & 98 & 98 & 98.04 \\
 & \multirow{4}{*}{\rotatebox[origin=c]{90}{176}} & R & 93.25 & 93.25 & 96.39 & 96.88 & 97.14 & 97.51 & 97.77 & 97.7 & 97.67 & 97.97 & 98.06 & 98.13 \\
 &  & P(50k) & 93.55 & 94.81 & 96.28 & 96.81 & 96.93 & 97.39 & 97.4 & 97.75 & 97.9 & 97.9 & 97.81 & 97.91 \\
 &  & P(100k) & 82.97 & 94.61 & 95.73 & 96.46 & 97.04 & 97.31 & 97.55 & 97.54 & 97.69 & 97.85 & 97.83 & 97.95 \\
 &  & P(200k) & 92.36 & 95.83 & 96.66 & 97.39 & 97.45 & 97.45 & 97.82 & 97.82 & 98.01 & 97.99 & 98.11 & 98.12 \\
 & \multirow{4}{*}{\rotatebox[origin=c]{90}{96}} & R & 62.53 & 76.4 & 86.28 & 90.83 & 94.25 & 94.25 & 95.86 & 95.86 & 95.86 & 96.21 & 96.49 & 96.78 \\
 &  & P(50k) & 90.5 & 90.5 & 93.87 & 93.87 & 94.5 & 95.57 & 95.96 & 95.96 & 95.96 & 96.82 & 96.82 & 96.82 \\
 &  & P(100k) & 70.31 & 91.09 & 94.87 & 94.73 & 94.73 & 94.57 & 96.18 & 96.45 & 96.45 & 96.77 & 96.77 & 96.77 \\
 &  & P(200k) & 87.83 & 91.26 & 93.18 & 93.58 & 94.86 & 95.28 & 96.02 & 96.02 & 95.98 & 95.98 & 95.98 & 95.98 \\ \hline
\multirow{12}{*}{\rotatebox[origin=c]{90}{U-Net 2 blocks}} & \multirow{4}{*}{\rotatebox[origin=c]{90}{116}} & R & 72.73 & 81.99 & 87.09 & 91.99 & 93.76 & 93.76 & 94.35 & 94.93 & 95.17 & 95.39 & 95.48 & 96.27 \\
 &  & P(50k) & 70.81 & 70.81 & 91.02 & 91.02 & 91.57 & 91.57 & 91.57 & 92.67 & 92.67 & 95.32 & 95.32 & 95.32 \\
 &  & P(100k) & 85.85 & 89.29 & 89.79 & 90.89 & 91.73 & 94.34 & 94.34 & 95.14 & 95.14 & 95.87 & 95.71 & 95.78 \\
 &  & P(200k) & 33.71 & 78.82 & 87.84 & 89.83 & 92.54 & 93.56 & 93.93 & 94.73 & 95.05 & 95.72 & 95.72 & 96.14 \\
 & \multirow{4}{*}{\rotatebox[origin=c]{90}{84}} & R & 71.23 & 66.65 & 88.39 & 88.39 & 91.5 & 91.5 & 92.9 & 93.42 & 94.1 & 94.07 & 92.15 & 92.15 \\
 &  & P(50k) & 68.8 & 68.8 & 78.39 & 78.39 & 88.28 & 88.28 & 88.28 & 90.64 & 90.64 & 90.1 & 90.1 & 93.04 \\
 &  & P(100k) & 83.48 & 84.8 & 84.28 & 89.55 & 91.64 & 91.64 & 91.64 & 92.63 & 93.23 & 92.04 & 92.04 & 92.04 \\
 &  & P(200k) & 78.24 & 78.24 & 85.15 & 87.72 & 89.22 & 89.22 & 89.22 & 90.36 & 90.74 & 91.94 & 91.94 & 91.94 \\
 & \multirow{4}{*}{\rotatebox[origin=c]{90}{84}} & R & 35.85 & 35.85 & 61.46 & 74.07 & 74.07 & 74.07 & 81.69 & 81.69 & 83.14 & 84.11 & 83.34 & 86.63 \\
 &  & P(50k) & 60.74 & 60.74 & 60.74 & 60.74 & 74.61 & 74.61 & 78.49 & 78.49 & 79.22 & 81.94 & 81.94 & 84.28 \\
 &  & P(100k) & 56.68 & 56.68 & 62.47 & 69.56 & 69.56 & 69.56 & 69.56 & 77.4 & 77.4 & 80.89 & 81.49 & 82.2 \\
 &  & P(200k) & 32.64 & 67.93 & 73.62 & 74.74 & 74.74 & 77.25 & 77.25 & 77.25 & 80.08 & 83.64 & 83.64 & 83.64
\end{tabular}%
}
\end{sc}
\end{small}
\end{center}
\vskip -0.1in
\end{table*}

\newpage
\subsection{Quantitative Evaluation on TEMImageNet datasets}
Similar to experiments on semantic segmentation, during training, the best model parameters are saved to disk after every 5 epochs. The following tables show model performance on the test datasets using the best models at the 5 epoch intervals. As observed in all tables, the higher performance is achieved in earlier epochs with fine-tuning while the random weight initialization leads to inferior performance or similar performance only towards the end of training.
\begin{table*}[!htb]
\caption{Comparison of the $L_1$ metric for different training methods (randomly initialized weights (R) and pretrained (P) with GANs on CEM500K using \SI{50}{K}, \SI{100}{K}, \SI{200}{K} images) on each of the downstream tasks: Super-resolution (SR), Noise \& Background Removal (N\&BGR) and Denoising (DN). The experiments were performed with HRNet.}
\label{au-table}
\vskip 0.15in
\begin{center}
\begin{small}
\begin{sc}
\resizebox{\textwidth}{!}{
\begin{tabular}{llllllllllllll}
\multicolumn{2}{l|}{epochs} & \multicolumn{1}{c}{5} & \multicolumn{1}{c}{10} & \multicolumn{1}{c}{15} & \multicolumn{1}{c}{20} & \multicolumn{1}{c}{25} & \multicolumn{1}{c}{30} & \multicolumn{1}{c}{35} & \multicolumn{1}{c}{40} & \multicolumn{1}{c}{45} & \multicolumn{1}{c}{50} & \multicolumn{1}{c}{55} & \multicolumn{1}{c}{60} \\ \hline
\multirow{4}{*}{SR} & \multicolumn{1}{l|}{R} & 0.01972 & 0.01731 & 0.01469 & 0.01636 & 0.01513 & 0.01524 & 0.01338 & 0.01208 & 0.01212 & 0.0113 & 0.01244 & 0.01122 \\
 & \multicolumn{1}{l|}{P(50k)} & 0.01688 & 0.01557 & 0.0177 & 0.01373 & 0.0139 & 0.01359 & 0.01239 & 0.01307 & 0.01137 & 0.01113 & 0.01155 & 0.0112 \\
 & \multicolumn{1}{l|}{P(100k)} & 0.02105 & 0.01685 & 0.01645 & 0.01451 & 0.0188 & 0.01465 & 0.01368 & 0.01357 & 0.01234 & 0.01231 & 0.01229 & 0.01142 \\
 & \multicolumn{1}{l|}{P(200k)} & 0.01639 & 0.01617 & 0.01544 & 0.01503 & 0.01428 & 0.01371 & 0.01249 & 0.01186 & 0.01222 & 0.01246 & 0.01159 & 0.01156 \\ \hline
\multirow{4}{*}{N\&BGR} & \multicolumn{1}{l|}{R} & 0.06391 & 0.04049 & 0.03073 & 0.0432 & 0.03312 & 0.03438 & 0.02974 & 0.03098 & 0.02102 & 0.01918 & 0.01905 & 0.0176 \\
 & \multicolumn{1}{l|}{P(50k)} & 0.03332 & 0.02916 & 0.03479 & 0.02494 & 0.03002 & 0.02257 & 0.02877 & 0.02326 & 0.0196 & 0.02056 & 0.01812 & 0.01776 \\
 & \multicolumn{1}{l|}{P(100k)} & 0.03268 & 0.04262 & 0.02986 & 0.03289 & 0.02772 & 0.02913 & 0.02307 & 0.02077 & 0.02126 & 0.01904 & 0.01973 & 0.01766 \\
 & \multicolumn{1}{l|}{P(200k)} & 0.03323 & 0.03545 & 0.02541 & 0.03602 & 0.02424 & 0.0224 & 0.02861 & 0.02136 & 0.02139 & 0.01851 & 0.01867 & 0.01792 \\ \hline
\multirow{4}{*}{DN} & \multicolumn{1}{l|}{R} & 0.04732 & 0.02965 & 0.03173 & 0.02433 & 0.02658 & 0.02996 & 0.03735 & 0.02042 & 0.02012 & 0.01982 & 0.01714 & 0.01541 \\
 & \multicolumn{1}{l|}{P(50k)} & 0.03287 & 0.0263 & 0.02193 & 0.02918 & 0.0212 & 0.02169 & 0.02778 & 0.01932 & 0.01727 & 0.01672 & 0.01613 & 0.01542 \\
 & \multicolumn{1}{l|}{P(100k)} & 0.03693 & 0.02867 & 0.02607 & 0.02381 & 0.01953 & 0.03708 & 0.02121 & 0.02148 & 0.01798 & 0.01751 & 0.01635 & 0.01555 \\
 & \multicolumn{1}{l|}{P(200k)} & 0.02657 & 0.02408 & 0.0212 & 0.02536 & 0.02489 & 0.01907 & 0.02233 & 0.01782 & 0.01674 & 0.01695 & 0.01638 & 0.01586
\end{tabular}%
}
\end{sc}
\end{small}
\end{center}
\vskip -0.1in
\end{table*}

%

\end{document}